%% file: main.tex
\documentclass{article}

\usepackage{microtype}
\usepackage{graphicx}
\usepackage{subfigure}
\usepackage{booktabs} 

\usepackage{hyperref}

\usepackage{amsmath}
\usepackage{caption}
\usepackage{subcaption}
\usepackage{wrapfig}
\usepackage{multirow}
\usepackage{amssymb}
\usepackage{pifont}
\usepackage{enumitem}

\input{math_commands.tex}

\usepackage[accepted]{icml2024}

\usepackage{amsmath}
\usepackage{amssymb}
\usepackage{mathtools}
\usepackage{amsthm}

\usepackage[capitalize,noabbrev]{cleveref}

\theoremstyle{plain}

\theoremstyle{definition}

\theoremstyle{remark}

\usepackage[textsize=tiny]{todonotes}

\icmltitlerunning{Isometric Representation Learning for Disentangled Latent Space of Diffusion Models}

\begin{document}

\twocolumn[
\icmltitle{Isometric Representation Learning for\\ Disentangled Latent Space of Diffusion Models}



\icmlsetsymbol{equal}{*}

\begin{icmlauthorlist}
\icmlauthor{Jaehoon Hahm}{equal,gsds}
\icmlauthor{Junho Lee}{equal,gsds}
\icmlauthor{Sunghyun Kim}{gsds}
\icmlauthor{Joonseok Lee}{gsds,google}
\end{icmlauthorlist}

\icmlaffiliation{gsds}{Seoul National University, Seoul, Korea}
\icmlaffiliation{google}{Google Research, Mountain View, California, United States}

\icmlcorrespondingauthor{Joonseok Lee}{joonseok@snu.ac.kr}

\icmlkeywords{Machine Learning, ICML}

\vskip 0.3in
]



\printAffiliationsAndNotice{\icmlEqualContribution} 
\newcommand\joonseok[1]{\textcolor{blue}{#1}}
\newcommand\jaehoon[1]{\textcolor{magenta}{#1}}
\newcommand\junho[1]{\textcolor{orange}{#1}}
\newcommand\sunghyun[1]{\textcolor{violet}{#1}}
\newcommand\red[1]{\textcolor{red}{#1}}
\newcommand\orange[1]{\textcolor{orange}{#1}}
\newcommand\blue[1]{\textcolor{blue}{#1}}

\renewcommand{\cite}{\citep}

\begin{abstract}
The latent space of diffusion model mostly still remains unexplored, despite its great success and potential in the field of generative modeling. In fact, the latent space of existing diffusion models are entangled, with a distorted mapping from its latent space to image space. To tackle this problem, we present \textit{Isometric Diffusion}, equipping a diffusion model with a geometric regularizer to guide the model to learn a geometrically sound latent space. Our approach allows diffusion models to learn a more disentangled latent space, which enables smoother interpolation, more accurate inversion, and more precise control over attributes directly in the latent space. Extensive experiments illustrate advantages of the proposed method in image interpolation, image inversion, and linear editing.
\end{abstract}

\input{1_intro}

\input{2_prelim}

\input{3_method}

\input{4_exp}
\input{5_related}

\input{6_summary}

\clearpage

\section*{Acknowledgements}

This work was supported by the
New Faculty Startup Fund from Seoul National University, by Samsung Electronics Co., Ltd (IO230414-05943-01, RAJ0123ZZ-80SD), by Youlchon Foundation (Nongshim Corp.), and by National Research Foundation (NRF) grants (No.
2021H1D3A2A03038607/50\%, RS-2024-00336576/10\%, RS-
2023-00222663/5\%) and Institute for Information \& communication Technology Planning \& evaluation (IITP) grants (No. RS-2024-00353131/25\%, 2022-0-00264/10\%), funded by the government of Korea.

\section*{Software and Data}


Our source code is publicly available at
\texttt{https://github.com/isno0907/isodiff}.
Readers would be able to reproduce the reported results by running this code.
We describe the detailed experimental settings including hyperparameters and hardware environments we use in Sec.~\ref{sec:exp:setting} and \ref{sec:exp:ablation}.

\section*{Impact Statement}

This paper proposes a method to enhance the underlying latent space of diffusion models to ease the image or video editing, selectively adjusting certain aspects of them as intended.
Our work shares ethical issues of generative models that are currently known in research community; to name some, deep fake, fake news, malicious editing to manipulate evidence, and so on.
We believe our work does not significantly worsen these concerns in general, but a better disentangled latent semantic space with our approach might ease these abuse cases as well.
Also, other relevant ethical issues regarding potential discrimination caused by a biased dataset still remain the same with our approach, neither improving nor worsening ethical concerns in this aspect.
A collective effort within the entire research community and society will be important to keep generative models beneficial.



\bibliography{main}
\bibliographystyle{icml2024}

\newpage
\appendix
\onecolumn

\input{7_appendix}


\end{document}

%% file: math_commands.tex
\usepackage{amsmath,amsfonts,bm}

\def\eqref#1{equation~\ref{#1}}

\def\1{\bm{1}}

\def\vh{{\bm{h}}}

\def\vv{{\bm{v}}}

\def\vx{{\bm{x}}}
\def\vy{{\bm{y}}}
\def\vz{{\bm{z}}}

\def\mI{{\bm{I}}}
\def\mJ{{\bm{J}}}

\DeclareMathAlphabet{\mathsfit}{\encodingdefault}{\sfdefault}{m}{sl}
\SetMathAlphabet{\mathsfit}{bold}{\encodingdefault}{\sfdefault}{bx}{n}

\newcommand{\E}{\mathbb{E}}

\newcommand{\R}{\mathbb{R}}

\newcommand{\Var}{\mathrm{Var}}

\DeclareMathOperator*{\argmin}{arg\,min}

%% file: 1_intro.tex
\section{Introduction}
\label{sec:intro}

\begin{figure*}
    \vspace{-0.3cm}
    \centering
    \includegraphics[width=\linewidth]{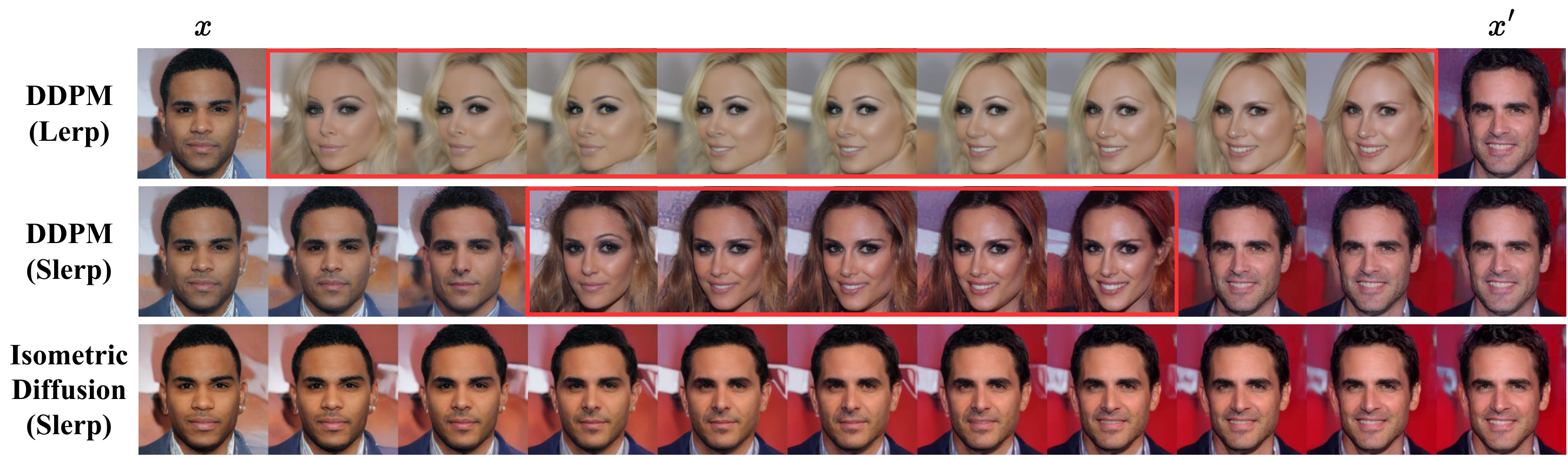}
    
    \vspace{-0.2cm}
    \caption{\textbf{An illustration of latent traversal between two latents $\vx$ and $\vx'$}.
    \textit{Top}: naive linear interpolation (Lerp) assuming Euclidean space, 
    \textit{Mid}: spherical interpolation (Slerp) between $\vx$ and $\vx'$ (direction $\vx \rightarrow \vx'$ is entangled with unwanted gender axis inducing abrupt changes),
    \textit{Bottom}: Slerp with the same latents with our Isometric Diffusion resolving unwanted entanglement.}

    \label{fig:video_interp}
    \vspace{-0.2cm}
\end{figure*}

Recently, diffusion models~\cite{sohl2015deep, song2019generative, ho2020ddpm, song2020score} have achieved unprecedented success across multiple fields, including image generation~\cite{dhariwal2021diffusion, nichol2021glide, ramesh2022hierarchical, saharia2022photorealistic, rombach2022high,lee2024posediff}, image editing~\cite{kawar2023imagic, ruiz2023dreambooth, hertz2022prompt}, video generation~\cite{ho2022imagen, blattmann2023align}, and scientific applications~\cite{cho2023hybrid}.
%
However, compared to other generative models like GANs~\cite{goodfellow2014generative} or VAEs~\cite{kingma2013auto}, there are few studies exploring the latent space of diffusion models. 

Learning a better latent space, particularly learning a \textit{disentangled} latent space has been historically an important problem in generative modeling.
The definition of a disentangled latent space varies depending on the field, but in generative modeling, it is defined as a latent space composed of linear subspaces, where each solely controls one factor of the variations \cite{bengio2013representation, higgins2017beta}.
Through various literatures on GANs~\cite{karras2020style2, chen2016infogan, shen2020interfacegan,kim2018disentangling} and VAEs~\cite{burgess2018beta, chen2018isolating}, disentanglement is known to be beneficial for downstream tasks such as image interpolation, inversion, and editing.
However, despite these benefits, only a few studies have addressed disentanglement of the latent space of diffusion models, possibly due to the relatively challenging analysis caused by their iterative sampling process.



Empirically exploring the latent space of diffusion models, we observe they are often entangled, aligned with recent discoveries and demonstration \cite{park2023understanding, peebles2022DiT}.
For example, a naive latent walking by linear interpolation between two latent vectors produces unwanted intermediate images, as illustrated in Fig.~\ref{fig:video_interp} (top).
Latent walking on a spherically interpolated trajectory between two latent vectors leads to a smoother intermediate images, as illustrated in
Fig.~\ref{fig:video_interp} (mid), but it is still not a geodesic on the data manifold; on the trajectory between two men, it unnecessarily goes through an unrelated woman.

This can be interpreted that there exist some distortions in the latent space of diffusion models, implying that they fail to accurately reflect the geometry of the data manifold; geodesic of latent space is not necessarily
mapped to geodesic on the data manifold.
Such a misalignment often leads to entanglement of multiple semantic concepts, which induces a sub-optimal image interpolation, image inversion, or
fine-grained image editing.

Motivated from the desire to guide diffusion models to learn a better disentangled latent space, we present \textit{Isometric Diffusion}, a diffusion model equipped with isometric representation learning.
Isometry is a map that preserves distance and angle between two metric spaces, and employing its geodesic preserving property, \textit{Isometric Diffusion} guides to obtain a geometrically sound latent space that better reflects the data manifold.
Specifically, we introduce a novel loss to encourage isometry between the latent space and the image space. With this additional guidance, latent walking induces a path closer to geodesic on the data manifold, and hence enables a smoother interpolation with less abrupt changes as in Fig.~\ref{fig:video_interp} (bottom).

To sum up, for the first time to the best of our knowledge, this paper proposes \textit{Isometric Diffusion}, a diffusion model equipped with geometric considerations that lead to a better disentangled latent space.
In order to obtain such geometrically sound latent space, we regularize the mapping from latent space to data manifold to be isometric.
Our proposed method achieves superior disentanglement, without substantial degradation in quality of the generated images.
We verify the effectiveness of our proposed method through quantitative and qualitative evaluations on various applications, including image interpolations, image inversions, and linear editing.

%% file: 2_prelim.tex
\section{Background}
\label{sec:els}

We briefly review the sampling and inversion techniques using DDIM \cite{song2020ddim}, latent spaces of diffusion models, and illustrate the objective for a better disentangled latent space.


\subsection{Diffusion Model}
\label{sec:els:dm}

\textbf{Training.}
Given an observed image space, denoted by $\mathcal{X}_0$, the forward process of diffusion models repeatedly perturbs an image $\vx_0 \in \mathcal{X}_0$ by $\vx_t = \sqrt{\bar{\alpha}_t} \vx_0 + \sqrt{1 - \bar{\alpha}_t} \mathbf{\epsilon}_0$,
with noise $\mathbf{\epsilon}_0 \sim \mathcal{N}(0, \mI)$ for $t = 1, ..., T$ where $\bar{\alpha}_t = \prod_{i=1}^{t}{\alpha_i}$.
These perturbed images $\vx_t$ construct a chain of latent spaces $\{\mathcal{X}_t\}$ for $t \in \{1, ..., T\}$, where the intermediate latent space at each time step $t$ is denoted by $\mathcal{X}_t$. 
For simplicity, we denote $\mathcal{X} \equiv \mathcal{X}_T$.
To recover the original image $\vx_0$ from $\vx_T$, diffusion models train a score model $\mathbf{s}_\theta$ by minimizing the following denoising score matching loss~\cite{vincent2011connection, song2020score}:
\begin{equation}
\resizebox{1.0\columnwidth}{!}{$
   \mathcal{L_\text{dsm}}(t) =~
   \lambda(t) \E_{\vx_0} \E_{\vx_t|\vx_0}
   \left[ \|\mathbf{s}_\theta(\vx_t, t) - \nabla_{\vx_t}\log p_t(\vx_t|\vx_0)\|_2^2 \right],
   $}
   \nonumber
   \label{eqn:training}
\end{equation}
where $\theta$ is a set of learnable parameters of the score model and $\lambda(t)$ is a positive weighting function.

\textbf{DDIM Sampling and Inversion.}
With the trained $\mathbf{s}_\theta$, we may generate an image $\vx_0$ from a sample $\vx_T \sim \mathcal{N}(0, \mI)$ through the reverse diffusion process. DDIM sampling accelerates the denoising process by skipping sampling steps \cite{song2020ddim, dhariwal2021diffusion}:
\begin{equation}
\resizebox{1.0\columnwidth}{!}{$
    \vx_{t-1} = \sqrt{\frac{\bar{\alpha}_{t-1}}{\bar{\alpha}_t}} \vx_t + \sqrt{\bar{\alpha}_{t-1}}\left(\sqrt{\frac{1}{\bar{\alpha}_{t-1}} - 1} - \sqrt{\frac{1}{\bar{\alpha}_{t}} - 1}\right) \epsilon_\theta(\vx_t, t).$}
    \label{eq:ddim_sampling}
    \nonumber
\end{equation}
DDIM inversion finds the corresponding latent of a given image $\vx_0$ by reversing the sampling process in the forward direction \cite{song2020ddim, dhariwal2021diffusion}:
\begin{equation}
\resizebox{1.0\columnwidth}{!}{$
    \vx_{t+1} = \sqrt{\frac{\bar{\alpha}_{t+1}}{\bar{\alpha}_t}} \vx_t + \sqrt{\bar{\alpha}_{t+1}}\left(\sqrt{\frac{1}{\bar{\alpha}_{t+1}} - 1} - \sqrt{\frac{1}{\bar{\alpha}_{t}} - 1}\right) \epsilon_\theta(\vx_t, t).$}
    \label{eq:ddim_inversion}
    \nonumber
\end{equation}

\subsection{Analysis on Latent Space $\mathcal{X}$ of Diffusion Models}
\label{sec:els:xspace}

The distribution of the norm of completely noised images $\|\vx_T\|_2$ follows a $\chi$-distribution, and they are distributed on the shell of a sphere, not uniformly within the sphere (see Sec.~\ref{sec:method:spherical} for more details).
For this reason, linearly interpolating two images within $\mathcal{X}$, as shown in Fig.~\ref{fig:video_interp} (top), results in a path far from geodesic on the data manifold, while spherical linear interpolation follows a shorter path.
However, as seen in Fig.~\ref{fig:video_interp} (mid), the spherical linear interpolation is still semantically not disentangled, indicating that entangled regions exist in $\mathcal{X}_T$.

\vspace{-0.1cm}
\subsection{Intermediate Latent Space $\mathcal{H}$ as a Semantic Space}
\label{sec:prelim:hspace}
\vspace{-0.1cm}

\citet{kwon2023diffusion} discovers that diffusion models have a semantic latent space $\mathcal{H}$ in the intermediate feature space of its score model. 
They suggest that the learned intermediate feature space $\mathcal{H}$ of the score model $\mathbf{s}_\theta$ sufficiently represents the semantics of the observed images.
Also, it is reported that a linear scaling by $\Delta \mathbf{h}$ on $\mathcal{H}$ controls the magnitude of semantic changes.

\subsection{Path Length Regularizer}
\label{sec:els:pl_reg}
Motivated to obtain a disentangled and smoother latent space of GANs, path length regularizer~\citep{karras2020style2} guides the generator $f: X \rightarrow Y$ to obtain a scaled-isometry, using an exponential moving average (EMA):
\begin{equation}
    \mathcal{L}_{\text{pl}}(f) = \E_{\vx,\vy \sim \mathcal{N}(0, \mI)} \left[ (\|{\mJ_\vx}^\top\vy\|_2 - a)^2 \right],
    \label{eq:plr}
\end{equation}
where $\vx \in X$ and $\vy \in Y$ are random samples from a normal distribution, $\mJ_\vx = \frac{\partial f}{\partial \vx}$ is the Jacobian of $f$, and $a$ is the exponential moving average of $\|{\mJ_\vx}^\top\vy\|_2$.
The objective is minimized when $\mJ_x$ is orthogonal up to a global scale. 

%% file: 3_method.tex
\section{Isometric Representation Learning for Diffusion Models}
\label{sec:method}

\begin{figure}
    \centering
    \includegraphics[width=\linewidth]{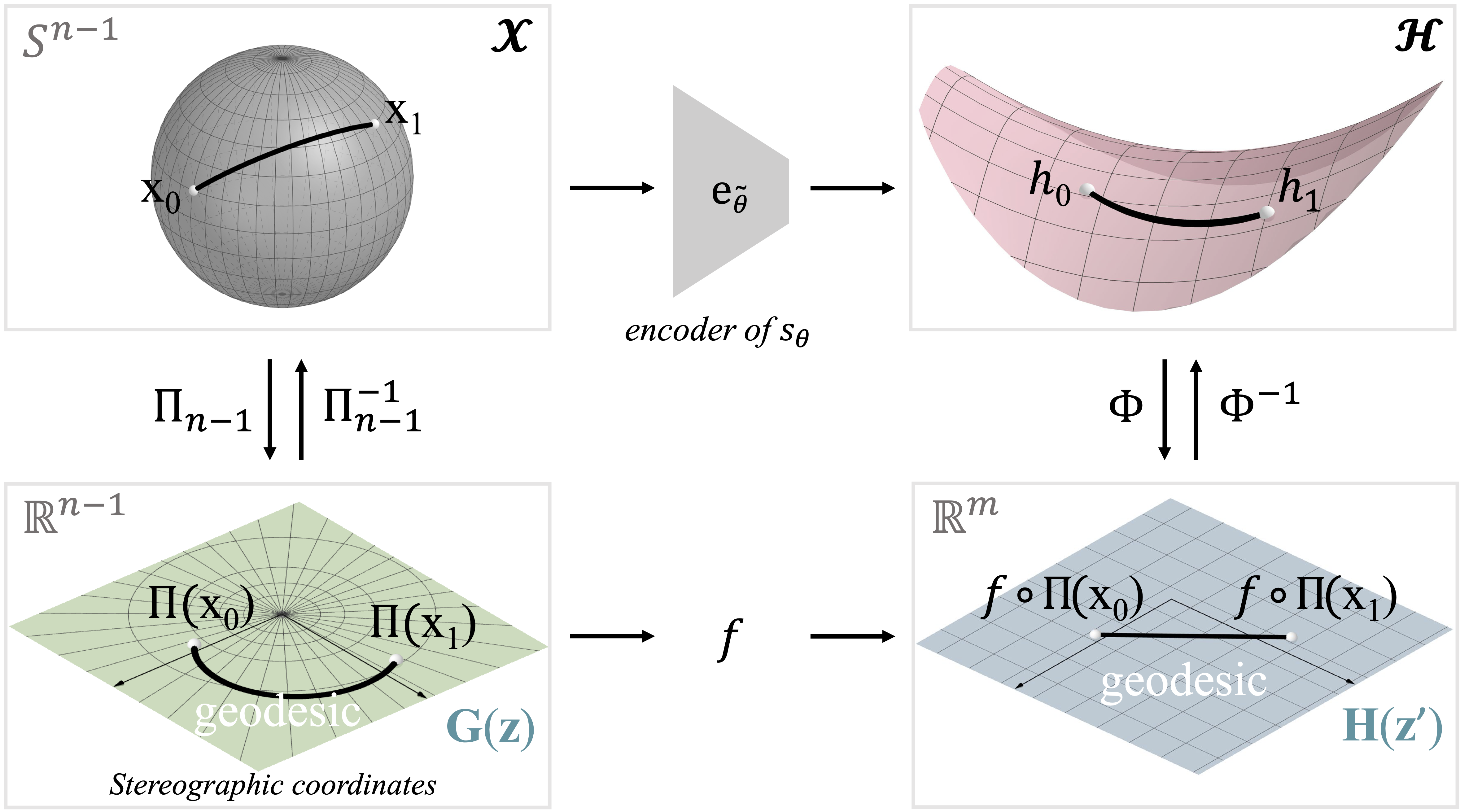}
    \caption{Illustration of $\mathcal{X}, \mathcal{H}$, and local coordinates of those two manifolds. Our isometric loss regularizes the encoder of the score model to map a spherical trajectory in $\mathcal{X}$ to a linear trajectory in $\mathcal{H}$, preserving a geodesic in $\mathcal{X}$ to a geodesic in $\mathcal{H}$. $e_{\tilde{\theta}}$ denotes the encoder of score model $s_\theta$. $\Pi_{n-1}$ and $\Phi$ are charts mapping from Riemmanian manifolds to local coordinate spaces. $\vz, \vz'$ denote the local coordinates of $\mathcal{X}, \mathcal{H}$, respectively.}
    \label{fig:overall}
\end{figure}

The goal of our work is to learn a latent space $\mathcal{X}$ which better reflects the geometry of the training data manifold by encouraging the mapping between them to be closer to geodesic-preserving.
We first explain the spherical approximation of latent space (Sec. \ref{sec:method:spherical}), definition and geodesic preserving property of scaled isometry (Sec. \ref{sec:method:isometric_map}), and how to guide the score model to learn an isometric mapping from $\mathcal{X}$ to $\mathcal{X}_0$ using a property of semantic latent space $\mathcal{H}$ (Sec. \ref{sec:method:isometry_loss}).
Fig.~\ref{fig:overall} illustrates the overall flow of our approach. 
Lastly, we discuss computational considerations (Sec.~\ref{sec:method:computation}).

\subsection{Spherical Approximation of the Latent Space}
\label{sec:method:spherical}





Recall that the sampling process of diffusion models starts from a Gaussian noise, $\vx_T \sim \mathcal{N}(0, \mI_n) \in \mathbb{R}^n$, where $T$ is the number of reverse time steps.
Then, the radii of Gaussian noise vectors $\vx_T$ follow $\chi$-distribution: $r = \sqrt{\sum_{i=1}^n \vx_{T,i}^2} \sim \chi(n)$, whose mean and variance are approximately $\sqrt{n-\frac{1}{2}} $ and 1, respectively.
For a sufficiently large $n$ (\emph{e.g.}, $n = 256 \times 256 \times 3$), the noise vectors reside within close proximity of a hypersphere with $r = \sqrt{n-\frac{1}{2}}$.





From this observation, we approximate the noise vectors $\vx \in \mathcal{X}$ (we omit subscripts to be uncluttered) reside on the hypersphere manifold $S^{n-1}(r) = \{ \vx \in \R^n: \|\vx\| = r\}$.
To define a Riemannian metric on $S^{n-1}(r)$, we need to choose charts and local coordinates to represent the Riemannian manifolds \cite{miranda1995algebraic}.
We choose the stereographic coordinates \cite{apostol1974mathematical} as the local coordinates to represent $\mathcal{X}$, and we set $\Phi = \text{id}$, the identity mapping defined at $\mathcal{H}$, which is the range of function we are interested in. 
Stereographic projection $\Pi_{n-1}: S^{n-1}(r) \setminus \{N\} \rightarrow \R^{n-1}$ is a bijective transformation from every point except for the north pole ($N$) on the hypersphere to a plane with the north pole as the reference point.
$\Pi_{n-1}$ and its inverse projection $\Pi_{n-1}^{-1}$ are given by

{ 
\vspace{-0.7cm}
\begin{align}
    &\Pi_{n-1}(\vx) = \frac{1}{r-\vx_n}(\vx_1, \vx_2, \cdots, \vx_{n-1}), \\
    &\Pi_{n-1}^{-1}(\vz) = \frac{r}{|\vz|^2 + 1}(2\vz_1, 2\vz_2, \cdots, 2\vz_{n-1}, |\vz|^2 - 1). \nonumber
\end{align}}

\vspace{-1em} \noindent
In stereographic coordinates, the Riemannian metric of the $S^{n-1}(r)$ \cite{do1992riemannian} is given by
\begin{equation}
    \mathbf{G_s}(\vz) = \frac{4r^4}{(|\vz|^2 + r^2)^2} \mI_{n-1}, \quad \forall \vz \in \R^{n-1}.
    \label{eq:gz}
\end{equation}
Recall that a diffusion model consists of a chain of latent spaces.
Hence, it is needed to verify at every time step the validity of spherical approximation.
From $\vx_t = \sqrt{\bar{\alpha}_t} \vx_0 + \sqrt{1 - \bar{\alpha}_t} \mathbf{\epsilon}_0$, the variance of perturbation kernels is $\Var[p(\vx_t|\vx_0)] = 1 - \bar{\alpha}_t = 1 - e^{\int -\beta(t) dt}$ \cite{song2020score}.
We use a linear noise schedule $\beta_t = \beta_0(1- \frac{t}{T}) + \beta_T \frac{t}{T}$ with $\beta_t = 1 - \alpha_t$.
We claim that for a sufficiently large $t$, $\sqrt{1 - \bar{\alpha}_t} \approx 1$ and thus the latent space can be approximated to a sphere.
That is, we approximate $\mathcal{X}_t \approx S^{n-1}(r)$ with $r=\sqrt{1 - \bar{\alpha}_t} \E[\chi(n)] = \sqrt{(1 - \bar{\alpha}_t)n}$ for $t > pT$, where we set $p \in [0, 1]$ as a hyperparamter.

\subsection{Isometric Mappings}
\label{sec:method:isometric_map}


\textbf{Definition.}
A mapping between two Riemannian manifolds $\mathbf{f}: \mathcal{M}_1 \rightarrow \mathcal{M}_2$ ({$f$ in local coordinates; $f = \Phi \circ {e_\theta} \circ \Pi_{n-1}^{-1}$}) is a \emph{scaled isometry}~\cite{yonghyeon2021irae} if and only if 
\begin{equation}
    \mathbf{G}(\vz) = c \mathbf{J}_f(\vz)^\top \mathbf{H}(f(\vz)) \mathbf{J}_f(\vz), \quad \forall \vz \in \mathbb{R}^{n-1},
\end{equation}
where $c \in \R$ is a constant, $\mathbf{J}_f(\vz) = \frac{\partial f}{\partial \vz} \in \R^{(n-1) \times m}$ is the Jacobian of $f$, $\mathbf{G}(\vz) \in \R^{(n-1) \times (n-1)}$ and $\mathbf{H}(\vz') \in \R^{m \times m}$ are the Riemannian metrics defined at the {local coordinates $\vz, \vz'$ of $\mathcal{M}_1 = \mathbb{R}^{n-1}$ and $\mathcal{M}_2 = \mathbb{R}^m$, respectively}.

Equivalently,
$f$ is a scaled isometry if and only if $\mathbf{J}_f^\top \mathbf{H} \mathbf{J}_f \mathbf{G}^{-1} = c\mI$ where $c \in \R$ is a global constant.
As its special case, $f$ is called a strict isometry when $c=1$, where a transformation between two metric spaces globally preserves distances and angles.
Scaled isometry allows the constant $c$ to vary, preserving only the \textit{scaled} distances and angles.
This relaxation makes it easier to optimize a function to preserve geodesic with less restrictions, hence leading to easier and more stable training than strict isometry.

In our problem formulation, $\mathcal{M}_1 = S^{n-1}$ ($\mathcal{X}$), $\mathcal{M}_2 = \R^m$ ($\mathcal{H}$), and $\mathbf{H}(\vz') = \mI_m$, as introduced in Sec.~\ref{sec:method:spherical}.
Although evaluation of $\mathbf{J}_f^\top \mathbf{H} \mathbf{J}_f \mathbf{G}^{-1}$ is coordinate-invariant, 
our choice of stereographic coordinates is computationally advantageous, as its Riemannian metric in Eq.~(\ref{eq:gz}) is proportional to the identity matrix (see Sec.~\ref{sec:method:computation} for details).

\textbf{Properties.}
To motivate the use of isometric mapping to learn disentangled representation, we introduce two important properties that isometry satisfies: geodesic-preserving and angle-preserving.
We follow the definition of disentanglement from \citet{bengio2013representation} and \citet{higgins2017beta}, which argue that a disentangled representation can be defined as one where a single latent unit is sensitive solely to changes in a single generative factor, while being invariant to changes in other factors.

\textit{1) Geodesic-preserving Property.}
Distance-preserving property of isometry naturally guarantees geodesic-preserving:

\vspace{-1em} \noindent
{\footnotesize \begin{align}
    &\argmin_{\gamma(t)} \int_0^1 \sqrt{\dot{\gamma}(t)^\top \mathbf{G} (\gamma(t)) \dot{\gamma}(t) } \mathrm{d}t \\
    &=
    \argmin_{\gamma(t)} \int_0^1 \sqrt{\dot{\gamma}(t)^\top \mathbf{J}(\gamma(t))^\top \mathbf{H}(f(\gamma(t))) \mathbf{J}(\gamma(t)) \dot{\gamma}(t) } \mathrm{d}t, \nonumber
\end{align}}

\vspace{-1em} \noindent
for an arbitrary trajectory $\gamma : [0,1] \rightarrow \R^n$ in local coordinates of $\mathcal{M}_1$ with fixed endpoints ($\gamma(0) = \vx_0, \gamma(1) = \vx_1$), where $\vx_0, \vx_1 \in \R^n$ are constant vectors and $\dot{\gamma}(t) = \frac{d\gamma}{dt}(t)$. 

This property induces equal sensitivity of each latent basis vector; a fixed-size step in the latent space results in equal amount of change in the semantic space, which is related to obtaining a smooth latent space.

\textit{2) Angle-preserving Property.}
This follows from the fact that if $G(x) = c J^\top(x) H(f(x)) J(x)$, then
\begin{align}
    \cos(\theta_1) 
    &= \frac{\langle v_1, v_2 \rangle_{\mathcal{M}_1}}{\|v_1\|_{\mathcal{M}_1} \|v_2\|_{\mathcal{M}_1} } \nonumber \\
    &= \frac{\langle df_p(v_1), df_p(v_2) \rangle_{\mathcal{M}_2}}{\|df_p(v_1)\|_{\mathcal{M}_2} \|df_p(v_2)\|_{\mathcal{M}_2}} = \cos(\theta_2),
\end{align}
where $\langle v_1, v_2 \rangle_{\mathcal{M}_1} = \dot{x_1}(0)^\top G \dot{x_2}(0)$, $\langle df_p(v_1), df_p(v_2) \rangle_{\mathcal{M}_2}$ 
$ = \dot{y_1}(0)^\top H \dot{y_2}(0) = \dot{x_1}(0)^\top J^\top H J \dot{x_2}(0)$, and $df_p$ is the pushforward at $p$.
$x_1(t), x_2(t), y_1(t), y_2(t)$ are the trajectories on manifolds $\mathcal{M}_1, \mathcal{M}_2$ such that $x_1(0)=p$, $x_2(0)=p$, $y_1(0)=f(p)$, $y_2(0) = f(p)$, and $\dot{x} = \frac{dx}{dt}(t)$.

Recalling the semantic space $\mathcal{H}$ discovered by \citet{kwon2023diffusion}, we pose that an orthogonal basis corresponding to meaningful visual attributes exists in the semantic space.
Due to the angle-preserving property, if the latent space $\mathcal{X}$ is mapped to $\mathcal{H}$ with an isometry, there exists orthogonal basis of $\mathcal{X}$ which is mapped to an orthogonal basis of $\mathcal{H}$ (assuming existence of the inverse).
This implies that a vector corresponding to a specific attribute is mapped to a single latent vector, orthogonal to other latent vectors corresponding to other factors.
This is related to the desired property of a disentangled latent space.

\subsection{Isometry Loss for Diffusion Models}
\label{sec:method:isometry_loss}

\textbf{Isometry Loss.}
To sum up, we can encourage the mapping $\mathbf{f}: \mathcal{X} \rightarrow \mathcal{H}$ to preserve geodesics and angles by regularizing $\mathbf{R}(\vz) \equiv \mathbf{J}_f(\vz)^\top \mathbf{H}(f(\vz)) \mathbf{J}_f(\vz) \mathbf{G}^{-1}(\vz) = c\mI$, for some $c \in \R$.
It can be achieved by minimizing the following isometry loss~\cite{yonghyeon2021irae}:
\begin{align}
    \mathcal{L}_\text{iso}(f, t) 
    &= \frac{ \E_{\vx_t \sim P(\vx_t)} [\mathrm{Tr}(\mathbf{R}^2(\vz_t))] }{\E_{\vx_t \sim P(\vx_t)} [\mathrm{Tr}(\mathbf{R}(\vz_t))]^2}
    \label{eq:loss_iso} \\
    &= \frac{ \E_{\vx_t \sim P(\vx_t)} \E_{\vv \sim \mathcal{N}(0,\mI)} [\vv^\top \mathbf{R}(\vz_t)^\top \mathbf{R}(\vz_t) \vv] }{\E_{\vx_t \sim P(\vx_t)} \E_{\vv \sim \mathcal{N}(0,\mI)} [\vv^\top \mathbf{R}(\vz_t) \vv]^2 },
    \label{eq:loss_iso} \nonumber
\end{align}
where $P(\vx_t)$ is the noise probability distribution at timestep $t$, and $\vz_t = \Pi_{n-1}(\vx_t)$.
The second equality holds due to the stochastic trace estimator~\cite{hutchinson1989trace}, where $\vv \in \R^{n-1}$ is a random vector such that $\E[\vv\vv^\top] = \mI$.

\textbf{Applying to Diffusion Models.}
Applying the isometry regularizer directly to the generating path of diffusion models is intractable, due to its iterative nature of sample generation.
Specifically, calculating $\mathbf{R}(\vz)$ in Eq. (\ref{eq:loss_iso}) requires the Jacobian of $f = f_0 \circ \cdots \circ f_{N-1}$, where $f_i$ is the $i$-th reverse step and $N$ is the number of reverse steps, resulting in a long chain of function compositions.

Motivated from the training method of diffusion models, we apply isometric regularizer at each time step.
To guide a mapping from $\mathcal{X}_T$ to $\mathcal{X}_0$ to be geodesic-preserving, we regularize each timestep of the iterative sequence; that is, the mapping between $\mathcal{X}_t$ and $\mathcal{X}_{t-1}$ for all $t \in \{T, ... ,1\}$.
Instead of regularizing all steps, we may selectively apply it.
For time steps closer to $T$, samples are closer to a Gaussian, so our assumption may reasonably hold.
For time steps closer to 0, samples are not sufficiently perturbed yet and thus they would follow some intermediate distribution between the Gaussian and the original data distribution.
Therefore, applying isometry loss to all timesteps can be sub-optimal and we let the portion of timesteps to apply it as a hyperparameter.

\begin{figure*}
    \centering
      \subfigure[\small{$S^2$ manifold}]{\includegraphics[width=0.23\textwidth]{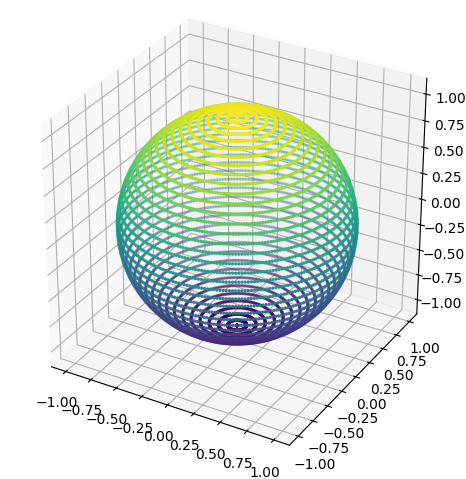}}
      \subfigure[\small{Reconstruction only}]{\includegraphics[width=0.23\textwidth]{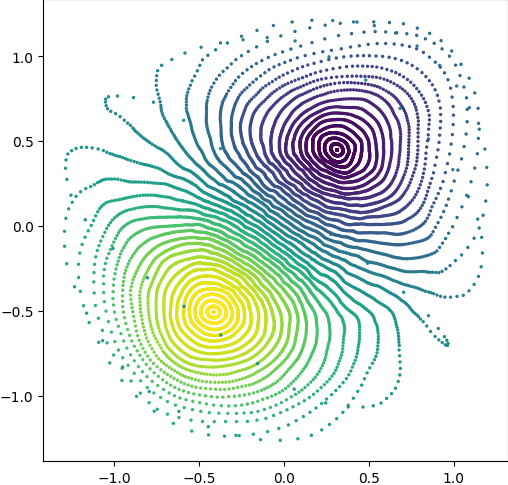}}
      \subfigure[$\mathcal{L}_\text{iso}$ with $\mathbf{G} = \mI$]{\includegraphics[width=0.23\textwidth]{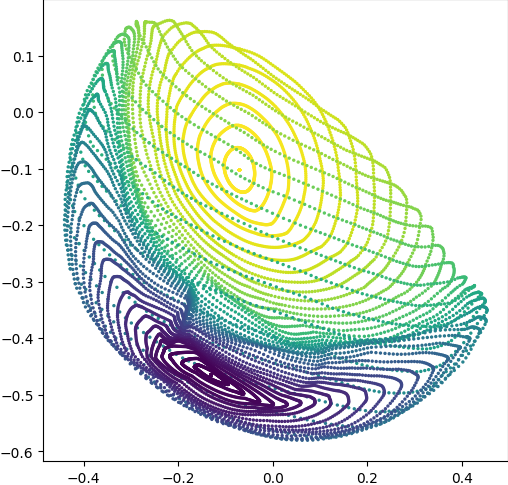}}
      \subfigure[$\mathcal{L}_\text{iso}$ with $\mathbf{G} = \mathbf{G_s}$ (Eq. (\ref{eq:loss_iso}))]{\includegraphics[width=0.23\textwidth]{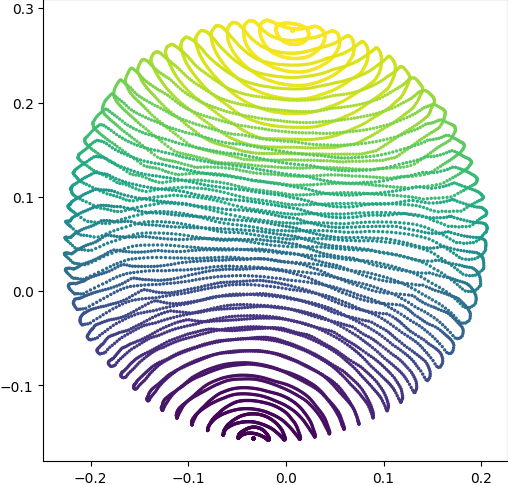}}
      \vspace{-0.2cm}
    \caption{(a) The input $S^2$ manifold. (b--d) Mapped contours in latent coordinates learned by an autoencoder; (b) with reconstruction loss only, (c) with isometric loss assuming navie Euclidean geometry, and (d) with our isometric loss considering $S^2$ geometry.}
    \label{fig:rm}
    \vspace{-0.2cm}
\end{figure*}

Also, to address the entanglement problem, we need to consider the semantic space of images rather than the pixel space.
Hence, we assume the semantic gap between images as a distance metric on $\mathcal{X}_T$.
The desired objective can be achieved by guiding the encoder of the score model, or equivalently a mapping from $\mathcal{X}_t$ to $\mathcal{H}_{t}$, to be more isometric.
Thus, we let $\mathbf{f} = e_{\tilde{\theta}}$, where $e_{\tilde{\theta}}$ denotes the encoder of score model $s_\theta$, and apply the isometry loss in Eq.~\eqref{eq:loss_iso}.




Our overall loss to train the score model is given by
\begin{equation}
    \mathcal{L}(t) = \mathcal{L}_\text{dsm}(t) +\lambda_\text{iso}(\gamma,t)\mathcal{L}_\text{iso}(e_{\tilde{\theta}}, t),
\end{equation}
where $\lambda_\text{iso}(p,t)$ is a non-negative weighting function 
and $\gamma \in [0, 1]$ is the ratio of timesteps to skip $\mathcal{L}_\text{iso}$.
That is, $\lambda_\text{iso}(\gamma,t) = \lambda_\text{iso}\1_{t'>\gamma T}(t'=t)$ where $\1(\cdot)$ is the indicator function, and the denoising process starts from $t=T$.



\textbf{Comparison with Path Length Regularizer.}
Calculated with exponential moving average (EMA), the path length regularizer~\cite{karras2020style2} may not equally penalize two mappings equivalent up to a global scale; that is, it may not hold $\mathcal{L}_{\text{pl}}(f) = \mathcal{L}_{\text{pl}}(f')$ even if ${\mathbf{J}_f}^\top \mathbf{J}_f = c{\mathbf{J}_f'}^\top {\mathbf{J}_f'}$ holds for some $c \in \mathbb{R}^{+}$, potentially leading to sub-optimal training.
In contrast, our isometric regularizer is scale-free, and does not require EMA-based optimization.
Thus, isometric regularizer can be seen as a generalization of the path length regularizer, and helps  to find the optimal point achieving disentanglement without significant degradation in generation quality.
We empirically demonstrate this in Sec.~\ref{sec:exp:ppl}.

\textbf{Illustration.}
We illustrate the purpose of isometric representation learning with a toy autoencoder example, learning an encoding map from $S^2$ to $\R^2$.
The autoencoder is trained with the reconstruction loss, regularized by our isometric loss in Eq. (\ref{eq:loss_iso}).
Fig.~\ref{fig:rm} illustrates an autoencoder flattening the given $S^2$ manifold in (a) with three different losses.
Only with the reconstruction loss, we see that the manifold in (b) is significantly distorted,
often locating two far-away points in the input closely in the latent space.
We observe clearly less distortion with the isometric loss in (c), under the assumption of the Euclidean metric in local coordinates of $S^2$ ($\mathbf{G} = \mathbf{I}$), but it still does not perfectly preserve geodesic.
With our full loss in (d), we see that the geometry of the input space is better preserved with $\mathbf{G} = \mathbf{G}_s$ from Eq. (\ref{eq:gz}).
We provide more illustrations in Appendix~\ref{appendix:illustration}.

\vspace{-0.1cm}
\subsection{Computational Considerations}
\label{sec:method:computation}
\vspace{-0.1cm}


To sidestep the heavy computation of full Jacobian matrices, we use stochastic trace estimator to substitute the trace of Jacobian to Jacobian-vector product (JVP).
Exploiting the commutativity and symmetry of the Riemmanian metric in stereographic coordinates, we utilize
{\scriptsize $\E_{\vv \sim \mathcal{N}(0,I)}[\vv^\top \mathbf{J}^\top \mathbf{J}\mathbf{G}^{-1}\vv] = \E_{\vv \sim \mathcal{N}(0, \mI)}[\vv^\top \sqrt{\mathbf{G}^{-\top}} \mathbf{J}^\top \mathbf{J}\sqrt{\mathbf{G}^{-1}}\vv] = \E_{\vv \sim \mathcal{N}(0, \mI)}[( \mathbf{J}\sqrt{\mathbf{G}^{-1}}\vv)^\top \mathbf{J}\sqrt{\mathbf{G}^{-1}}\vv]$}
to reduce the number of JVP evaluations.
We provide more details about the computation of stochastic trace estimator in Appendix~\ref{appendix:ste:computation}.

%% file: 4_exp.tex
\section{Experiments}
\label{sec:exp}

We conduct extensive experiments to verify the effectiveness of our isometric loss $\mathcal{L}_\text{iso}$ on disentangling the latent space of diffusion models.
We obtain experimental results by fine-tuning a pre-trained model with our $\mathcal{L}_\text{iso}$, unless noted otherwise.
Refer to Appendix. \ref{appendix:implementation} for further details.


\begin{table*}
    \caption{\textbf{Quantitative comparison.} Diffusion models trained with our isometry loss achieve consistent improvement over the baselines.}
    \vspace{0.1cm}
    \centering
    \footnotesize
    \renewcommand{\tabcolsep}{6pt}
    \begin{tabular}{lc|rr|rr|rr|rr|rr|cc} 
        \toprule
            &
            & \multicolumn{2}{c|}{FID-10k$\downarrow$}
            & \multicolumn{2}{c|}{PPL-50k$\downarrow$}
            & \multicolumn{2}{c|}{mRTL$\downarrow$}  
            & \multicolumn{2}{c|}{MCN $\downarrow$}
            & \multicolumn{2}{c|}{VoR $\downarrow$} 
            & \multicolumn{2}{c}{LS $\downarrow$}
            \\
    
        
            \multicolumn{1}{l}{Dataset}
            & \multicolumn{1}{l|}{Model}
            & \multicolumn{1}{c}{Base} & \multicolumn{1}{c|}{Ours}
            & \multicolumn{1}{c}{Base} & \multicolumn{1}{c|}{Ours}
            & \multicolumn{1}{c}{Base} & \multicolumn{1}{c|}{Ours} 
            & \multicolumn{1}{c}{Base} & \multicolumn{1}{c|}{Ours}
            & \multicolumn{1}{c}{Base} & \multicolumn{1}{c|}{Ours}
            & \multicolumn{1}{c}{Base} & \multicolumn{1}{c}{Ours} 
            \\ 
    
        \midrule
        
        \multirow{1}{*}{CIFAR-10} & DDPM  
            & \textbf{10.19}   & 10.50 
            & 126   & \textbf{101}   
            & 2.03 & \textbf{1.92}
            &  155       &  \textbf{107}
            &  \textbf{0.50}  &  0.57
            & -             & -
            \\

        \multirow{1}{*}{LSUN-Church} & DDPM
            &  \textbf{10.56}  &  12.10 
            &    2028 &   \textbf{1559}
            &   3.71  &   \textbf{3.21}    
            &   375      &  \textbf{217}
            &   1.92      &  \textbf{1.37}
            &   -    & -
            \\
    
        \multirow{1}{*}{LSUN-Bedrooms} & DDPM
            &  \textbf{11.95}  &   12.02
            &          4515   &   \textbf{3809} 
            &  3.38         &  \textbf{3.21}  
            &  320       &   \textbf{186}
            &  1.69    &  \textbf{1.12}
            &  -     & -
            \\
    
        \multirow{1}{*}{CelebA-HQ} & DDPM
            &  \textbf{15.89} &  16.18 
            &  648     &  \textbf{455} 
            &  2.67    &   \textbf{2.50} 
            &  497     & \textbf{180}
            &  1.42    & \textbf{0.85}
            &  1.91     & \textbf{1.51}
            \\

        \midrule
        
            
        \multirow{1}{*}{CelebA-HQ} & LDM
            &  \textbf{10.79} &  11.46
            &  439     &  \textbf{397} 
            &  2.89    &   \textbf{2.73} 
            &  322     & \textbf{198}
            &  1.04    & \textbf{0.54}
            &  2.38             & \textbf{2.15}
            \\
        \bottomrule    
    \end{tabular}
    
    \label{table:fid_various}
    \vspace{-0.2cm}
\end{table*}

\subsection{Experimental Settings}
\label{sec:exp:setting}

\textbf{Dataset.}
We evaluate our approach on CIFAR-10, CelebA-HQ~\cite{huang2018introvae}, 
LSUN-Church and LSUN-Bedrooms~\cite{wang2017knowledge}.
The training partition of each dataset consists of 50K, 14K, 126K, and 3M samples, respectively.
We resize each image to $256 \times 256$ except for CIFAR-10 and horizontally flip it with probability 0.5.

\textbf{Evaluation Metrics.}
\emph{Fréchet inception distance (FID)} \cite{heusel2017fid} is a widely-used metric to assess the quality of images created by a generative model by comparing the distribution of generated images with that of ground truth images.
\emph{Perceptual Path Length (PPL)}~\cite{karras2019style} evaluates how well the generator interpolates between points in the latent space, defined as $\text{PPL} = \E [\frac{1}{\epsilon^2}d(\vx_t, \vx_{t+\epsilon})]$, where $d(\cdot, \cdot)$ is a distance function.
We use LPIPS~\cite{zhang2018lpips} distance using AlexNet \cite{krizhevsky2012alexnet} for $d$.
A lower PPL indicates a better disentangled latent space, since when two or more axes are entangled and geodesic interpolation in $\mathcal{X}$ induces a sub-optimal trajectory in the semantic space, the LPIPS distance gets larger and thereby so does the PPL.
We perform 20 and 100 steps of DDIM sampling for FID and PPL, computed with 10,000 and 50,000 images, respectively.
\emph{Linear separability (LS)} \cite{karras2019style} measures the degree of entanglement of a latent space, by measuring how much the latent space is far from being separable by a hyperplane.
Since LS requires attributes, we measure it only on CelebA-HQ.
\emph{Mean condition number} (MCN) and \emph{variance of Riemannian metric} (VoR) measure how close a mapping is to a scaled-isometry, proposed by \citet{yonghyeon2021irae}.
We provide further details on these metrics in Appendix~\ref{appendix:metrics}.

We additionally design a new metric called \emph{mean Relative Trajectory Length (mRTL)}, measuring the extent to which a trajectory in $\mathcal{X}$ is mapped to geodesic in $\mathcal{H}$.
Specifically, mRTL is defined as the mean ratio between the $L_2$ distance $d_{2}(t)$ of $\vh, \vh' \in \mathcal{H}$, corresponding to two latents $\vx, \vx' \in \mathcal{X}$, and another distance measured on the manifold $d_\mathcal{M}(t)$, following the mapped path on $\{\mathcal{H}_t\}$.
That is, $\text{RTL}(t) = \E_{\vx, \vx' \in \mathcal{X}} \left[d_\mathcal{M}(t) / d_{2}(t) \right]$ and $\text{mRTL} = \E_t[\text{RTL}(t)]$, where $t$ denotes the timesteps of the sampling schedule.
Intuitively, it represents the degree of isometry of the encoder $\mathbf{f}$.





\subsection{Quantitative Comparison}
\label{sec:exp:ppl}

\textbf{Overall Comparison.}
We quantitatively evaluate the effect of our method on DDPM~\cite{ho2020ddpm} and unconditional latent diffusion model (LDM) \cite{rombach2022high} on various datasets.
Tab.~\ref{table:fid_various} indicates that the diffusion models trained with our isometric regularizer exhibit substantial improvement in PPL, implying smoother transitions during latent traversal.
Smaller mRTL, MCN, and VoR also signify that the encoder of score model gets closer to scaled-isometry with our method.
On CelebA-HQ, LS significantly drops, indicating improved disentanglement of the latent space.



\textbf{FID and Disentanglement Trade-off.}
As shown in Tab.~\ref{table:fid_various}, applying our regularizer appears to introduce some trade-off between FID and the disentanglement metrics.
However, we emphasize that low FID and nice disentanglement are two distinct desired aspects of image generation tasks, and their importance may vary depending on the user's needs.
For instance, let us assume that a generator $f_{\text{gen}}$ has learned the exact distribution of the training dataset $\mathcal{D}$, $p_\text{data}(y) = \frac{1}{|\mathcal{D}|} \sum_{y_i \in \mathcal{D}}{\delta(y - y_i)}$, where $\delta(\cdot)$ denotes Dirac-delta function.
That is, $f_{\text{gen}}(x) = y_i$ if $x \in X_i$, where $\{X_i\}$ is a partition of $\mathrm{dom}(f_{\text{gen}})$, indicating a mode-collapsed generator.
In this case, it would achieve the lowest FID, but this is not a desired generative model, as it would result a maximal entanglement in the latent space.
Consequently, it could be evaluated as a poor generator for downstream tasks such as inversion, image editing, and interpolation.
Our proposed method provides a systematic way for the users to efficiently adjust the relevant importance of these two aspects by setting the regularization coefficient $\lambda_\text{iso}$, according to their needs depending on the specific target task.

Additionally, \citet{karras2020style2} discovers correlation between the perceived image quality and PPL metric.
They explain that FID cannot fully characterize the generation quality of a generative model and demonstrate qualitative comparisons, claiming that lower PPL with the same FID relates to higher image quality.
This shows achieving a low PPL is also relevant to high quality of generated images.


\input{tables/plr_vs_iso}

\begin{figure}
  \centering
    \includegraphics[width=0.82\linewidth]{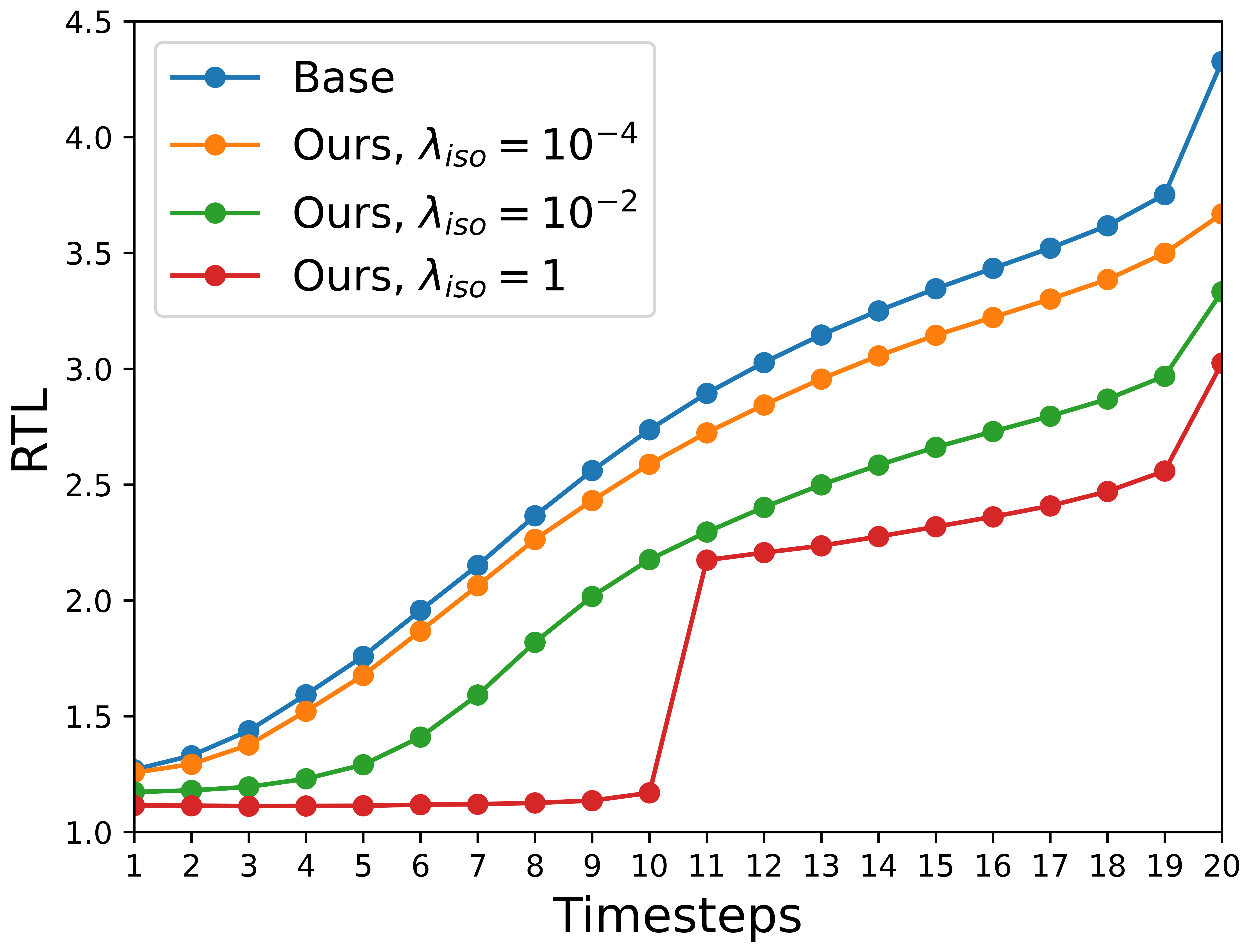}
    \vspace{-0.2cm}
    \caption{\textbf{RTL with various $\lambda_{\text{iso}}$.} A stronger regularization reduces the ratio to 1, flattening the trajectories in $\mathcal{H}$.}
    \label{fig:pathlength}
    \vspace{-0.1cm}
\end{figure}

\textbf{Comparison with Path Length Regularizer.}
As mentioned in Sec.~\ref{sec:els:pl_reg}, EMA training of path length regularizer $\mathcal{L}_\text{pl}$ can be sub-optimal,
while isometric regularizer $\mathcal{L}_\text{iso}$ is scale-free.
Indeed, from Tab.~\ref{table:plr_vs_iso}, we observe that using $\mathcal{L}_\text{pl}$ slightly improves PPL from the baseline while significantly worsens FID.
On the other hand, regularizing via $\mathcal{L}_\text{iso}$ with $\mathbf{G_s}$, considering the accurate geometry of the latent space, significantly improves PPL while maintaining FID.
Also, as seen in Tab.~\ref{table:inversion} and Fig.~\ref{fig:inversion}, our method demonstrates superior performance in inversion and reconstruction downstream tasks.
These experiments demonstrate that our isometric regularizer makes the training more stable and easier.

\textbf{Mean Relative Trajectory Length.}
Fig.~\ref{fig:pathlength} shows the measured Relative Trajectory Length (RTL) scores across the reverse timesteps in DDIM ($T=20$).
As the guidance of isometric loss gets larger with a larger $\lambda_\text{iso}$, the RTL tends to decrease, indicating the geodesic in $\mathcal{X}$ (Slerp) maps to geodesic in $\{\mathcal{H}_t\}$.
We notice a significant drop when $t \leq 10$ especially with a larger $\lambda_\text{iso}$, where the isometric loss is applied.
This indeed shows the isometric loss is accurately guiding the encoder of the score model to learn an isometric representation.

\subsection{Analysis on the Disentanglement of Latent Space $\mathcal{X}$}
\label{sec:exp:analysis}

We demonstrate that the disentangled latent space obtained with our method is advantageous in various downstream tasks such as interpolation, inversion, and linear editing.

\begin{figure*}[t]
    \centering
    \includegraphics[width=\linewidth]{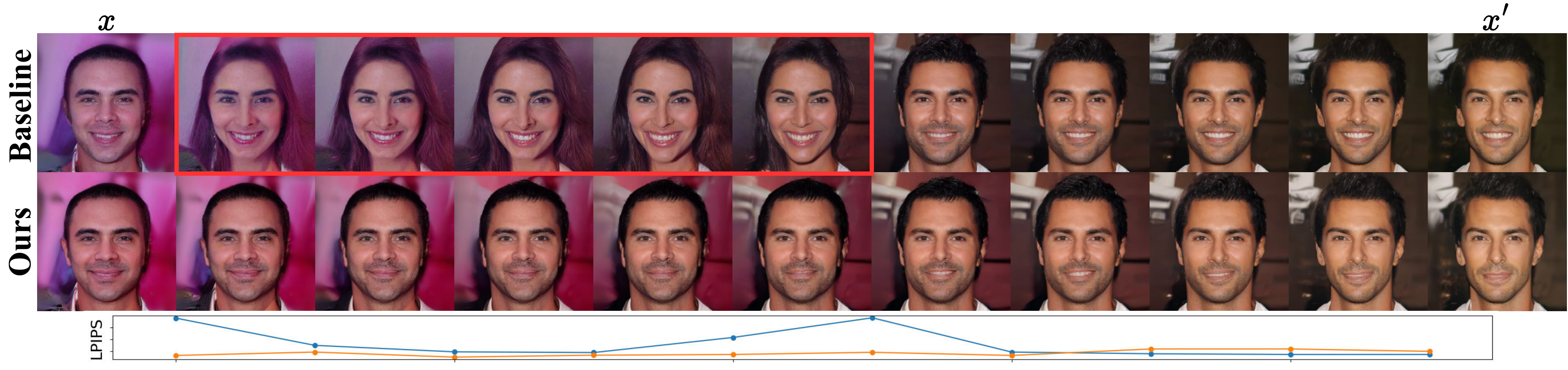}
    \vspace{-0.05cm}
    \includegraphics[width=\linewidth]{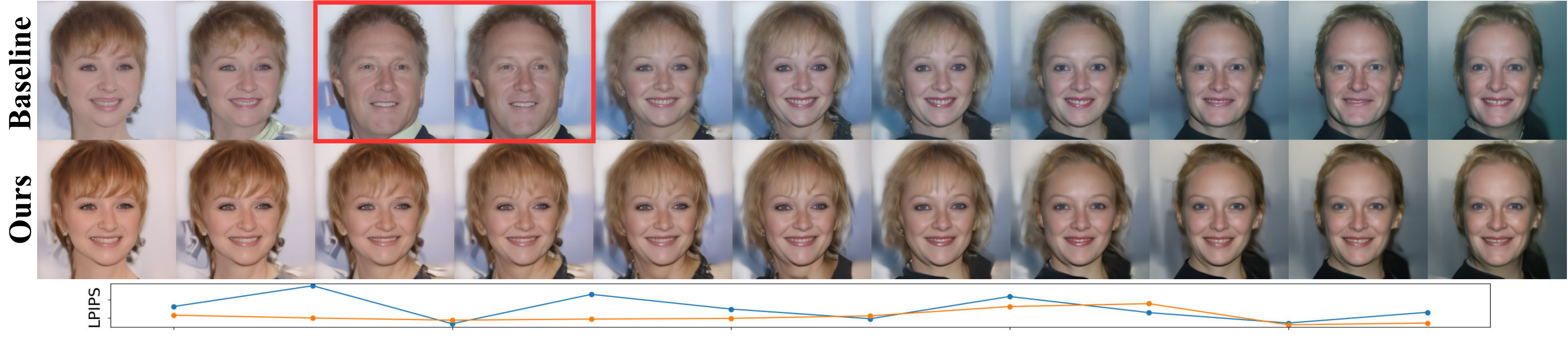}
    \includegraphics[width=\linewidth]{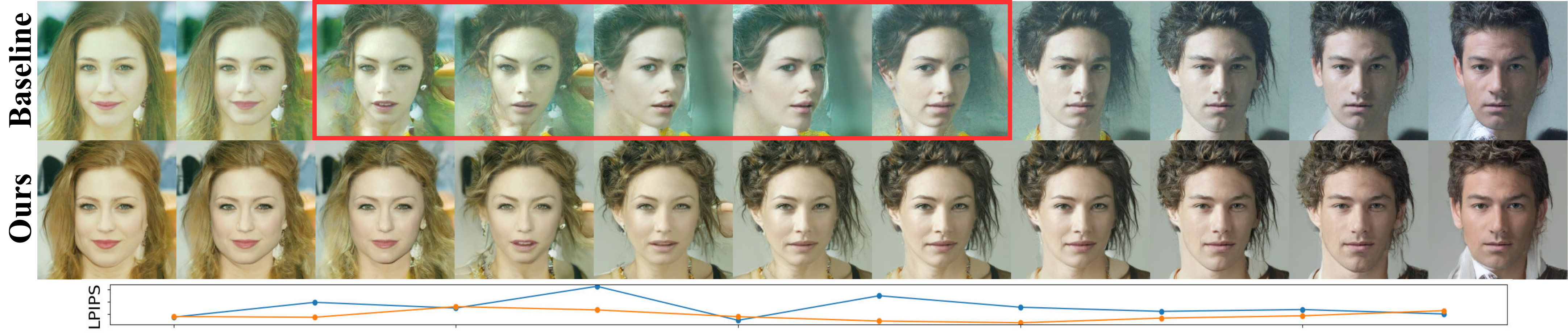}
    \vspace{-0.6cm}
    \caption{\textbf{Image interpolation.} {Examples of latent traversal between two latents $\vx$ and $\vx'$ with DDPM \cite{ho2020ddpm}, trained on $256 \times 256$ CelebA-HQ.
    We observe unnecessary changes of female $\rightarrow$ male in the baseline, while smoother transitions in ours. For quantitative support, we plot LPIPS distance between each adjacent frames (\blue{Blue}: Base, \orange{Orange}: Ours).} }
    \label{fig:interpolation}
    \vspace{-0.2cm}
\end{figure*}

\textbf{Interpolation.}
We first conduct traversals on the latent space $\mathcal{X}$ between two points $\vx, \vx' \in \mathcal{X}$, illustrating the generated images from interpolated points between them in Fig. ~\ref{fig:interpolation}.
We observe that with our isometric loss the latent space is better disentangled, resulting in smoother transitions without abrupt changes in gender.
More examples are provided in Fig.~\ref{fig:interpolation2} in Appendix~\ref{appendix:interpolation}.

\begin{table}
    \caption{\textbf{Quantitative comparisons of image inversion and reconstruction.} We employ DDIM inversion to convert source image to latent, and reconstruct the image with DDIM sampling. Note that low PPL relates to better inversion and reconstruction.}
    \hspace{1cm}
    \centering
    \renewcommand{\tabcolsep}{3pt}
    \resizebox{\columnwidth}{!}{
    \begin{tabular}{l|c|cccc} 
        \toprule
            Regularizer & PPL-50k
            & \multicolumn{1}{c}{MSE $\downarrow$} & \multicolumn{1}{c}{PSNR $\uparrow$} 
            & \multicolumn{1}{c}{SSIM $\uparrow$} & \multicolumn{1}{c}{LPIPS $\downarrow$}  \\ 
    
        \midrule

        { - } & 401  & 0.00862   & 0.597 & 20.6   & 0.517  \\
        
         {$\mathcal{L}_\text{pl}$ \text{(Path length reg.)}} & 368   & 0.00667   & 0.614  & 21.7   & 0.521  \\

        
        {$\mathcal{L}_\text{iso}$ \text{(Ours)}} & \textbf{340}  & \textbf{0.00599}    & \textbf{0.674}  & \textbf{22.2}   & \textbf{0.436}  \\
           
        \bottomrule    
    \end{tabular}
    }
    \label{table:inversion}
\end{table}

\textbf{Inversion and Reconstruction.}
In literature of GANs \cite{karras2020style2}, achieving a lower PPL and consequently having a disentangled latent space is beneficial for image inversion and reconstruction.
Achieving accurate inversion and reconstruction is particularly important for image editing with diffusion models because it consists of inverting the given image into a latent, and the editing happens in that latent space.
Thus, we conduct similar experiments on inversion and reconstruction on diffusion, using DDIM~\cite{song2020ddim} and ADM~\cite{dhariwal2021diffusion} trained on CelebA-HQ.

\begin{figure}[t]
    \centering
    \includegraphics[width=\linewidth]{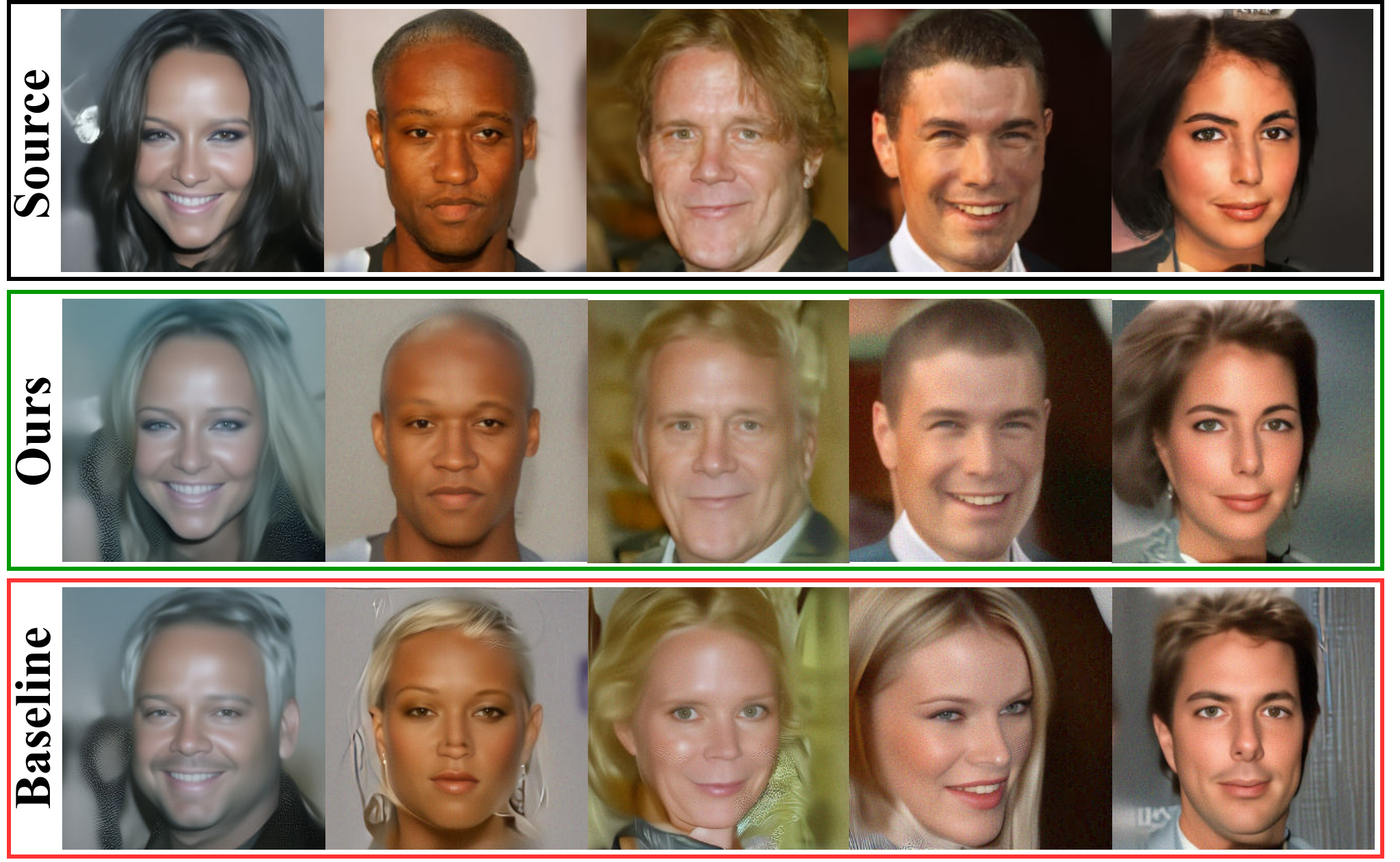}
    \vspace{-0.8cm}
    \caption{\textbf{Image inversion and reconstruction.} Baseline is ADM \cite{dhariwal2021diffusion} trained on 256$\times$256 CelebA-HQ.}
    \label{fig:inversion}
    \vspace{-0.3cm}
\end{figure}

Tab.~\ref{table:inversion} reports the effect of our method on the image inversion and reconstruction tasks.
Particularly, the PPL is a direct metric to measure disentanglement, and thus a lower PPL with our method strongly indicates better quality of image inversion.
Fig.~\ref{fig:inversion} qualitatively illustrates the advantage of our method in inversion and reconstruction.

\begin{figure*}
    \centering
    \vspace{-0.2cm}
    \includegraphics[width=0.9\linewidth]{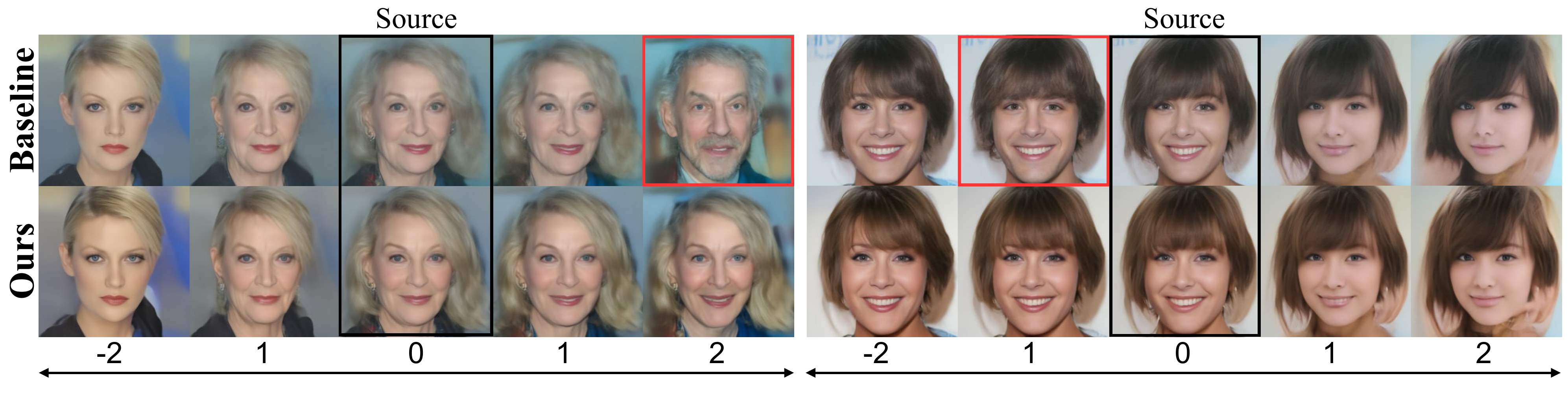}
    \vspace{-0.4cm}
    \caption{\textbf{Linearity.} Images generated from a source latent vector $\vx$ and from slightly perturbed latents, $\vx + \gamma\Delta \vx$ with $\gamma \in \{-2, -1, 0, 1, 2\}$, where $\Delta \vx$ corresponds to the change in age axis. 
    }
    \label{fig:linearity}
    \vspace{-0.6cm}
\end{figure*}

\textbf{Linearity.}
We also claim that the latent space $\mathcal{X}$ learned with our isometric loss has a property of \textit{linearity}.
Specifically, we compare the generated images with ours to baseline, where both are moved along the slerp in their latent spaces.
For this, we find the editable direction following~\citet{jang2022geometrically}, an unsupervised method for identifying semantic-factorizing directions in the latent space based on its local geometry, and perturb the latents through this direction both for baseline and our model.
In this way, we discover the principal variations of the latent space in the neighborhood of the base latent code.

Fig.~\ref{fig:linearity} demonstrates that a spherical perturbation on $\mathcal{X}$ with various intensity of $\Delta \vx$ adds or removes specific attributes from the generated images accordingly.
As seen in Fig.~\ref{fig:linearity}, the baseline often changes multiple factors (age, gender) abruptly and inconsistently with $\gamma$ (\emph{e.g.}, when $\gamma = -1$ on the right example, it suddenly shows a male-like output), while ours show disentangled changes.

\begin{figure*}[ht]
  \centering
  \vspace{0.2cm}
  \subfigure[DDPM (base)]{\includegraphics[width=0.4\textwidth]{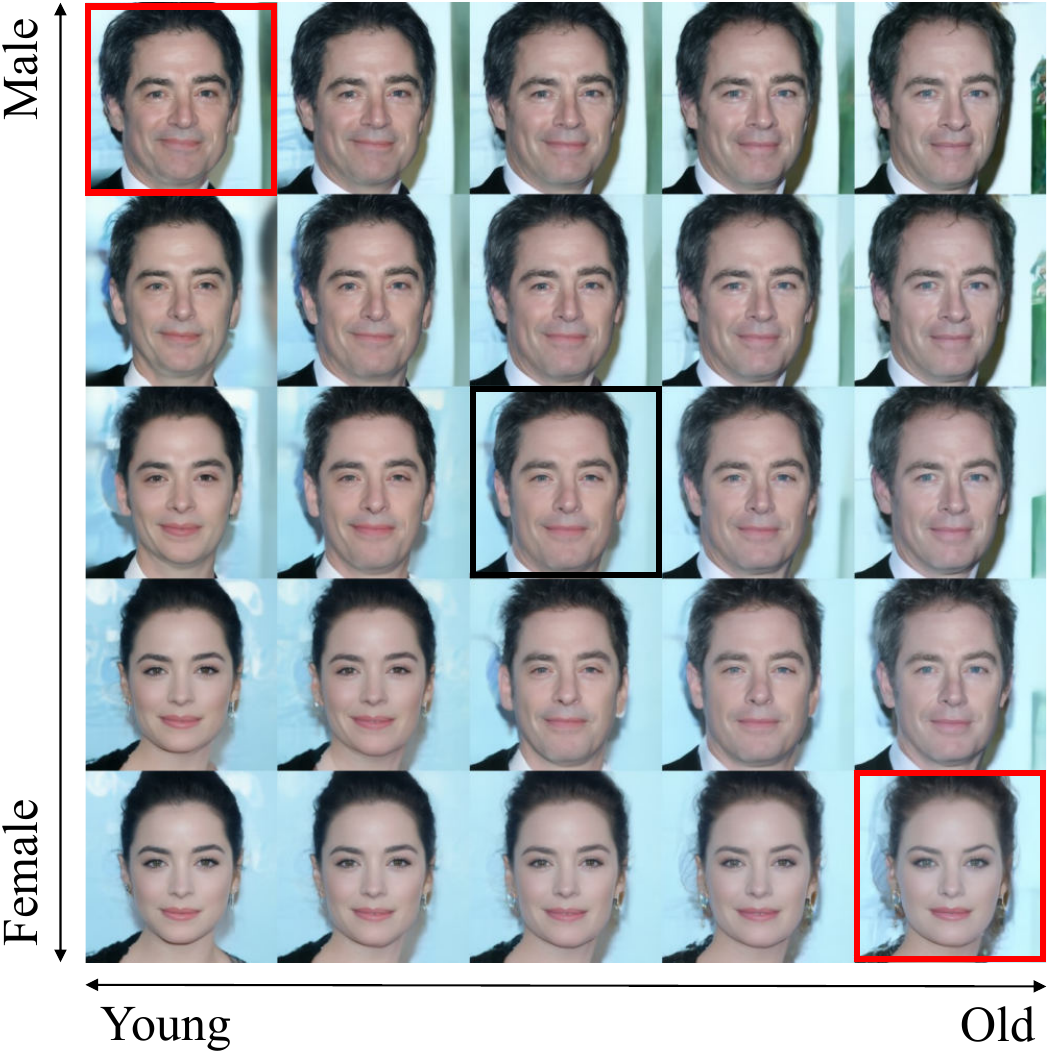}}
  \subfigure[\textit{Isometric Diffusion} (ours)]{\includegraphics[width=0.4\textwidth]{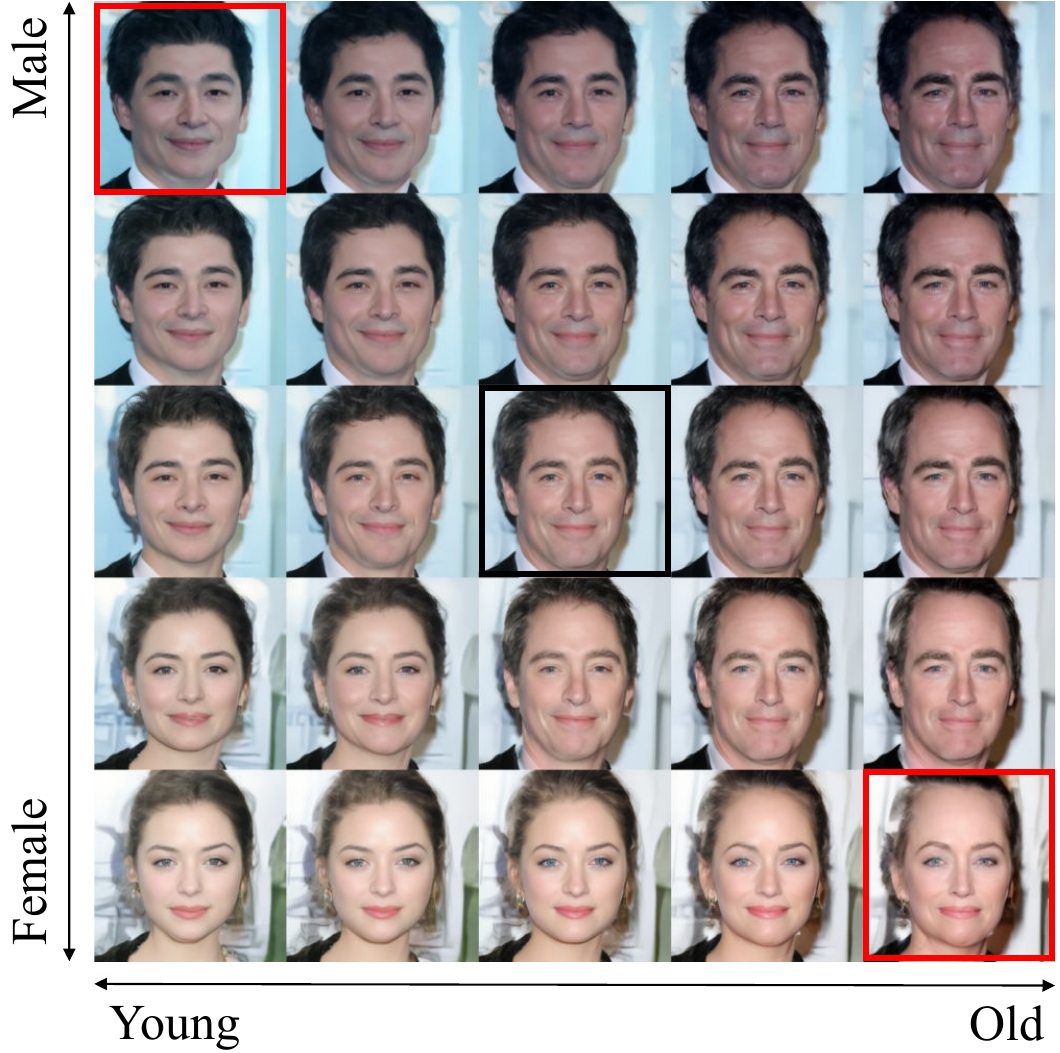}}
  \vspace{-0.2cm}
  \caption{\textbf{Linear combination in $\mathcal{X}$}. With ours, age and gender axes are better disentangled. Image generated with the source latent is marked with black box.}
  \vspace{-0.3cm}
  \label{fig:2d_grid}
\end{figure*}

Fig.~\ref{fig:2d_grid} further illustrates the linearity of $\mathcal{X}$ with images manipulated in two directions in $\mathcal{X}$.
For this, we follow \citet{choi2021not} to find the editing directions.
Comparing the results of baseline and ours, we observe that our method better disentangles the concept of age and gender, successfully drawing a young male and an old female (marked with red boxes), where the baseline fails to.
This indicates that the latent space trained with our approach is better disentangled, and they can be easily combined back with a linear combination.



\subsection{Ablation Study}
\label{sec:exp:ablation}

Tab. \ref{tab:ablation} shows the ablation study on the choice of optimal $p$ and $\mathbf{G}$.
We observe the best performance with $\gamma=0.5$ and $\mathbf{G} = \mathbf{G}_{s}$, in FID and PPL.
Note that $\gamma = 1$ denotes the original training of diffusion model.
Also, using a proper Riemannian metric $\mathbf{G}$ of the latent space when calculating the isometric loss turns out to be important.
This result supports our idea to model the latent space of diffusion model as a Riemannian manifold $S^{n-1}$ is indeed reasonable.

\input{tables/ablation}

%% file: tables/plr_vs_iso.tex
\begin{table}
    \vspace{-0.3cm}
    \caption{\textbf{Isometric  \textit{vs.} Path length regularizers.} Ours with the correct Riemannian metric ($\mathbf{G}$) leads to better FID and PPL. 
    }
    \vspace{0.1cm}
    \centering
    \small
    \begin{tabular}{l|c|cc} 
        \toprule        
        Regularizer & $\mathbf{G}$ & FID-10k$\downarrow$ & PPL-50k$\downarrow$ \\
        \midrule
        - & - & 15.89  & 648 \\
        $\mathcal{L}_{\text{pl}}$  (Path length reg.)  & $\mathbf{I}$ & 20.04  & 552 \\
        $\mathcal{L}_{\text{iso}}$ & $\mathbf{I}$ & 16.60  & 619 \\
        $\mathcal{L}_{\text{iso}}$ (Ours) & $\mathbf{G_s}$ & 16.18  & \textbf{455} \\
        \bottomrule    
    \end{tabular}
    \label{table:plr_vs_iso}
\end{table}

%% file: tables/ablation.tex
    
        


    
    

\begin{table}
    \vspace{-0.4cm}
    \caption{\textbf{Ablation study} on $\gamma$, the ratio of timesteps to skip applying isometric loss, and $\mathbf{G}$, the choice of Riemannian metric.     
    }
    \centering
    \small
    \begin{tabular}{ccc|cc}
        \toprule
         $\gamma$ & $\mathbf{G}$ & $\lambda_{\text{iso}}$ & FID-10k $\downarrow$ &  PPL-50k $\downarrow$ \\        
        \midrule

         1      & -            & -         & 15.89  & 653  \\
         0      & $\mathbf{I}$ & $10^{-4}$ & 24.07  & 447  \\
         0.5    & $\mathbf{I}$ & $10^{-3}$ & 30.28  & 441  \\
         0.5    & $\mathbf{I}$ & $10^{-4}$ & 16.60  & 619  \\

        \midrule

         0.5 & $\mathbf{G}_{s}$ & $10^{-4}$ & 16.18 & 455  \\

        \bottomrule
    \end{tabular}
    \vspace{-0.5cm}
    \label{tab:ablation}
\end{table}

    

%% file: 5_related.tex
\section{Related Work}
\label{sec:related}


\textbf{Latent Space of Generative Models.}
On Generative Adversarial Networks (GANs) \cite{goodfellow2014generative,radford2015unsupervised,zhu2017unpaired,choi2018stargan, ramesh2018spectral,harkonen2020ganspace,abdal2021styleflow}, StyleGAN \cite{karras2019style} is a pioneering work on latent space analysis and improvement.
In StyleGANv2 \cite{karras2020style2}, a path length regularizer guides the generator to learn an isometric mapping from the latent space to the image space.
Recently, additional studies on GANs \cite{shen2020interpreting,shen2020interfacegan,shen2021closed} and VAEs \cite{hadjeres2017glsr,zheng2019disentangling,zhou2020learning} have examined the latent spaces of generative models. 
\citet{kwon2023diffusion} found that the internal feature space of U-Net in diffusion models, $\mathcal{H}$, plays the same role as a semantic latent space. \citet{preechakul2022diffusion} discovered that using a semantic encoder enables the access to the semantic space of diffusion models. However, this method utilizes additional conditioning information, while our work proposes a method that can directly utilize the latent space without any condition.


\textbf{Riemannian Geometry for Generative Models.}
There exist some previous works on utilizing Riemannian geometry to understand the latent spaces.
\citet{chen2020fmvae} proposed that interpreting the latent space as Riemannian manifold and regularizing the Riemannian metric to be a scaled identity help VAEs learn a good latent representation.
\citet{yonghyeon2021irae} proposed an isometric regularization method for geometry-preserving latent space coordinates in scale-free and coordinate invariant form, arguing that an isometrically regularized autoencoder is advantageous in image retrieval task.
\citet{arvanitidis2021latent} claimed understanding Riemmanian geometry of the latent space and directly incorporating the pullback metric can improve analysis of representations as well as generative modeling.
However, this method can be computationally heavy.
Our method focuses on the reduction of computation cost at inference.
See Appendix.~\ref{appendix:pullback} for further discussions.

%% file: 6_summary.tex
\section{Summary}
\label{sec:conclusion}



In this work, we address a critical challenge in the field of generative models, particularly disentangling latent space for diffusion models. Despite the notable progress of diffusion models in generating photorealistic samples, there persists a substantial gap in comprehending and controlling their latent spaces.

Motivated from isometric representation learning, our \textit{Isometric Diffusion} introduces a novel regularizer aimed at obtaining a more disentangled latent space for diffusion models.
Through a mapping from latent space to data manifold being close to isometry, our approach demonstrates the attainment of a more intuitive and disentangled latent space for diffusion models, as evidenced both quantitatively and qualitatively.
We demonstrate advantages of achieving disentangled and smoother latent space through extensive experiments of image interpolation, inversion and linear editing.

Our method will open up new possibilities for practical applications, including video generation with seamless transitional frames and easier manipulation of specific features, providing a high degree of control and customization.
We believe our method can be applied to conditional generation, which will be a promising future work.




%% file: 7_appendix.tex
\clearpage

\appendix

\pagenumbering{roman}
\renewcommand\thetable{\Roman{table}}
\renewcommand\thefigure{\Roman{figure}}
\setcounter{table}{0}
\setcounter{figure}{0}

\section{Implementation Details}
\label{appendix:implementation}
Our network architecture follows the backbone of DDPM~\cite{ho2020ddpm}, which uses a U-Net~\cite{ronneberger2015unet} internally. If not specified, we train with batch size 32, learning rate $10^{-4}$, $p=0.5$, and $\lambda_\text{iso}=10^{-4}$ for 10 epochs by default.

For all datasets and models, we initialize with pre-trained weights and further fine-tune them with each competing method until the lowest FID is achieved.
All the scores reported in Tab.~\ref{table:fid_various} have been achieved before 5000 iterations.
We experimentally confirm that the results of training from the scratch and fine-tuned are almost identical.

We use Adam optimizer and exponential moving average~\cite{brown1956exponential} on model parameters with a decay factor of 0.9999.
We use 4 NVIDIA A100 GPUs with 40GB memory for experiments.

\section{Details on Evaluation Metrics}
\label{appendix:metrics}

In this section, we provide further details of the evaluation metrics we use throughout this paper.


\emph{Linear separability (LS)} \cite{karras2019style} measures the degree of disentanglement of a latent space.
\citet{karras2019style} argues that if a latent space is disentangled, it should be able to find a consistent direction that changes an image attribute independently, and thus the latent space labeled according to the specific attribute should be separable by a hyperplane.
The formal definition of this metric is as follows:
\begin{equation}
  \text{LS} = e^{\sum_i {H(Y_i|X_i)}},
\end{equation}
where $i$ is the attribute index, $H(\cdot|\cdot)$ is conditional entropy, $X$ are the classes predicted by SVM, and $Y$ are the classes predicted by a pre-trained classifier.
Intuitively, it measures how much additional information is needed to fully determine the label determined by the classifier, knowing the label predicted by SVM, hence indicating how much the latent space is separable by a hyperplane.

We train a classifier with ResNeXt \citep{xie2017aggregated} to predict the 40 attribute confidence scores with CelebA annotated for each image, and then follow the method in \cite{karras2019style}.
We calculate it with SVMs using linear kernel and radial basis function kernel, regarding the spherical geometry of the latent space.
We compute it with 1,000 images pruned after sorting with classifier confidence scores, from 2,000 images generated.

\emph{Mean condition number} (MCN) and \emph{variance of Riemannian metric} (VoR) are the metrics measuring how much a mapping is close to a scaled-isometry, proposed by \cite{yonghyeon2021irae}.
We measure MCN and VoR of the score models' encoders to measure how much our isometric regularizer has successfully guided the encoder to be isometric.
Formally, the mean condition number (MCN) is defined as
\begin{equation}
  \text{MCN} = \E_{\vx_0}\E_{\vx_t \sim p(\vx_t|\vx_0)} \left[ \frac{\sigma_{M}(\mJ(\vx_t))}{\sigma_{m}(\mJ(\vx_t))} \right],
\end{equation}
where $\sigma_{M}, \sigma_{m}$ are the maximum and minimum singular values.
MCN measures how isotropic the Riemannian metric is.
Note that $\sigma_i(\mJ(\vx_t)) = \lambda_i^2(\mJ^\top(\vx_t)\mJ(\vx_t))$, where $\lambda_i$ is the $i$-th eigenvalue.
The variance of Riemannian metric (VoR) is defined as
\begin{equation}
  \text{VoR} = \sum_i \Var_{\vx_0, \vx_t \sim p(\vx_t|\vx_0)} \left[ \sigma_i(\mJ(\vx_t)) \right],
\end{equation}
where we measure how homogeneous Riemannian metric is.
Note that we slightly modify its definition to bypass the exact calculation of Jacobian by exploiting SVD.
Satisfying both isotropicity and homogeneity of Riemannian metric, a mapping can be determined its proximity to isometry.
We measure them with 1,000 images.


\section{Advantages of Disentangled Latent Space}
\label{appendix:advantages}

While there exists some topological discrepancy between Gaussian prior and the true image distribution, generative modeling have often modeled their latent spaces as Gaussian (\textit{e.g.}, GANs, VAEs) and there have been studies on the advantages of geometric regularizing in learning a `better' latent space modeled as Gaussian, even though the target distribution will be quite different from it. We believe that such geodesic preserving property is motivated from various literatures in generative models.

For example, StyleGAN2~\cite{karras2020style2} uses path length regularizer to guide the generator to become closer to isometry and achieves a smoother latent space. Their work shows that the path-length-regularized StyleGAN2 improves 1) to lower PPL (a consistency and stability metric in image generation), and 2) to have invertibility from image to its latent codes. We believe the latter is potentially related to the existence of smooth inverse function of the generator, which is an important feature for image manipulation. In diffusion models, this corresponds to DDIM inversion~\cite{dhariwal2021diffusion}, and we believe our method can improve the inversion quality in diffusion models and hence contribute to high quality latent manipulations, with similar effects with that of path length regularized StyleGAN2.

Additionally, FMVAE~\cite{chen2020fmvae} uses isometric regularizer to the decoder of VAE to learn a mapping from Gaussian latent space to image space close to isometry, obtaining advantages in downstream tasks using geometrically aligned latent space. As also illustrated in~\citet{karras2020style2} and~\citet{chen2020fmvae}, we admit that it somehow penalizes the FID score, possibly due to the nature of regularizer. We leave the exploration of minimizing the tradeoff as a promising future work.


Also, disentangled latent space leads to improvement in image editing capabilities.
First of all, disentangled representations make image editing more effective and intuitive, since it becomes easier to manipulate specific attributes of an image without affecting others when the underlying key factors are disentangled.
For example, if a model has disentangled representations for pose and identity in images of faces, one could edit the pose of a face without altering its identity, or vice versa.
We demonstrate in Fig. \ref{fig:inversion} that the advantages of disentangled latent space in the inversion and reconstruction task, which is particularly important for image editing with diffusion models.
This is because image editing consists of inverting the given image into a latent, and the editing happens in that latent space.






\section{On the Scalability of the Proposed Method}
\label{appendix:scalability}

As discussed in \citet{park2023understanding}, the complexity of $\mathcal{H}$ increases as the complexity of the training dataset increases.
The work also explicitly reports the entanglement phenomena empirically discovered in Stable Diffusion~\cite{rombach2022high}, marking as its limitation.
While intervention of large-scale training data, latent encoder/decoder, and text encoder in latent diffusion models (LDM) or Stable Diffusion complicates the relation between the noise space ($\mathcal{X}$) and the semantic space ($\mathcal{H}$), \citet{jeong2023training} demonstrates the efficacy of $\mathcal{H}$ space also in Stable Diffusion in a text-conditioned setting, hence validating the method also in large-scale setting. 

Therefore, we believe the method can be scaled up, and also can incorporate conditional models including text-to-image models such as Stable Diffusion, which can be an interesting direction for future work. 
As long as the $\mathcal{H}$ space is effective, our approach can be easily adopted to further regularize it with minimal additional cost.


\section{On the Challenges of Directly Applying Pullback Metric to Diffusion Models}
\label{appendix:pullback}

Under the setting of using VAE in \citet{arvanitidis2021latent}, pulling back the metric of the observed space could be straightforward, since the generator is explicitly defined with VAE.
However, since the generative process of diffusion model is iterative, directly translating this method to diffusion models can be infeasible.
Specifically, pulling back the metric of the observed space requires calculating the Riemmanian metric $\mathbf{G} = \frac{\partial f}{\partial \vx}^\top \frac{\partial f}{\partial \vx}$ for every point on the interested trajectory.
This requires full calculation of Jacobian of $f = f_0 \circ \cdots \circ f_{N-1}$, where $N$ is the number of reverse steps (\textit{e.g.}, $N=100$ in DDIM) and $f_i$ is the $i$-th reverse step, resulting in a long chain of function compositions.
This could be computationally expensive for heavy models such as high resolution diffusion models.



Also, in order to obtain geodesic, one needs to numerically solve a corresponding ODE or to directly optimize discretized trajectory, and this additional step also can be computationally expensive.
Our method proposes to transfer this computation from inference time to training time, and this is beneficial in a sense that inference can be done many times while training will be done only once.

Furthermore, assuming calculation of the pull back metric in the diffusion model is feasible, directly utilizing the pullback metric and our method are not conflicting but complementary to each other.
Our approach improves the latent space but can take further benefit by direct methods like \citet{arvanitidis2021latent}, by obtaining exact geodesics and fully reflecting the geometry of observed space to the latent space.


\section{Stochastic Trace Estimator}
\label{appendix:ste}
\subsection{Estimation Accuracy}
\label{appendix:ste:acc}
In Eq.~(\ref{eq:loss_iso}) of the main text, we explained that the second quality holds because of the stochastic trace estimator~\cite{hutchinson1989trace} which is an algorithm to obtain such an estimate from matrix-vector products:
\begin{equation}    
    \mathrm{Tr}(A)=\E[\vv^\top A \vv] \simeq \frac{1}{N}\sum_{i=1}^{N}v^T_i A v_i,
\end{equation}
where $A$ is any square matrix and $\vv$ is random vector such that $\E[\vv\vv^\top] = \mI$. 

\begin{figure*}[ht]
    \centering
    \includegraphics[width=0.8\linewidth]{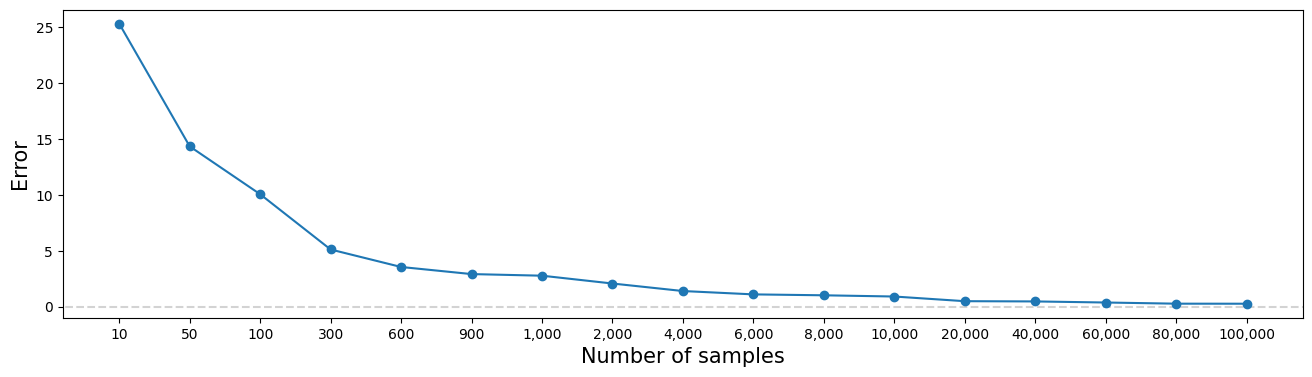}
    \caption{\textbf{Approximation error of stochastic trace estimator against the number of samples.} Each point on the graph represents the error corresponding to a particular sample size.}
    \label{fig:ste_error}
\end{figure*}

As shown in Fig. \ref{fig:ste_error} the error of stochastic trace estimator increases as the number of sample $N$. In this experiment, $A$ follows $\mathcal{N}(0, \mI) \in \mathbb{R}^{256 \times 256}$ and $\vv$ follows $\mathcal{N}(0, \mI) \in \mathbb{R}^{256 \times 1}$.

\begin{figure}
      \centering
      \subfigure[$S^2$ manifold]{\includegraphics[width=0.23\textwidth]{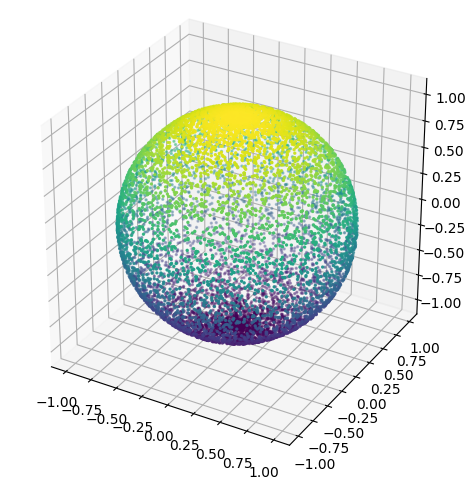}}
      \subfigure[$\mathcal{L}_\text{iso}$ with estimator]{\includegraphics[width=0.23\textwidth]{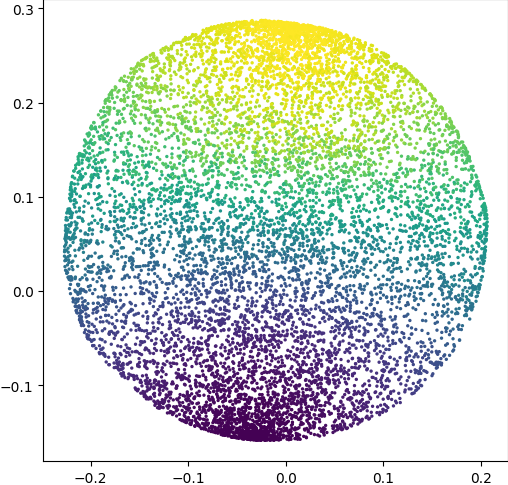}}
      \subfigure[$\mathcal{L}_\text{iso}$ with exact trace]{\includegraphics[width=0.23\textwidth]{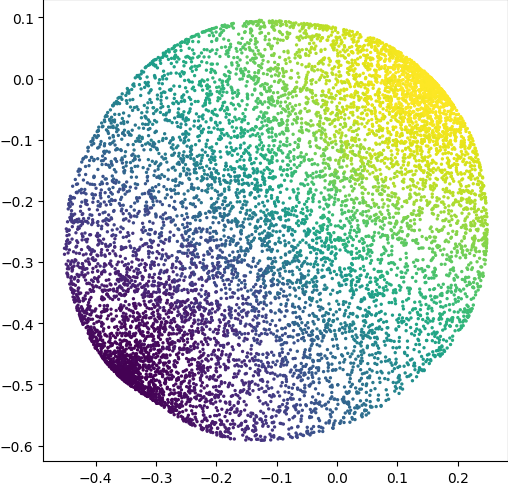}}
    \caption{(a) Illustration of the input $S^2$ manifold. (b) latent coordinates learned with isometric regularizer, estimated with the stochastic trace estimator. (c) latent coordinates learned with exact isometric regularizer. }
    \label{fig:ste_vs_exact}
\end{figure}

Despite the inherent errors of estimator, we conduct a simple experiment in the setting similar to Fig.~\ref{fig:rm} to investigate whether optimizing with estimated trace converges similar to optimizing with exact trace. As shown in Fig.~\ref{fig:ste_vs_exact}, optimizing the model by approximating the trace of the matrix with the stochastic trace estimator yields similar results to those obtained by using the actual trace of the matrix. Furthermore, Fig.~\ref{fig:ste_error} demonstrates that the approximated trace exhibits a similar convergence pattern in loss over training time. These results suggest that the final convergence point is similar even when the loss function is optimized by estimating the trace of the matrix through stochastic trace estimator.

\begin{figure*}[ht]
    \centering
    \includegraphics[width=\linewidth]{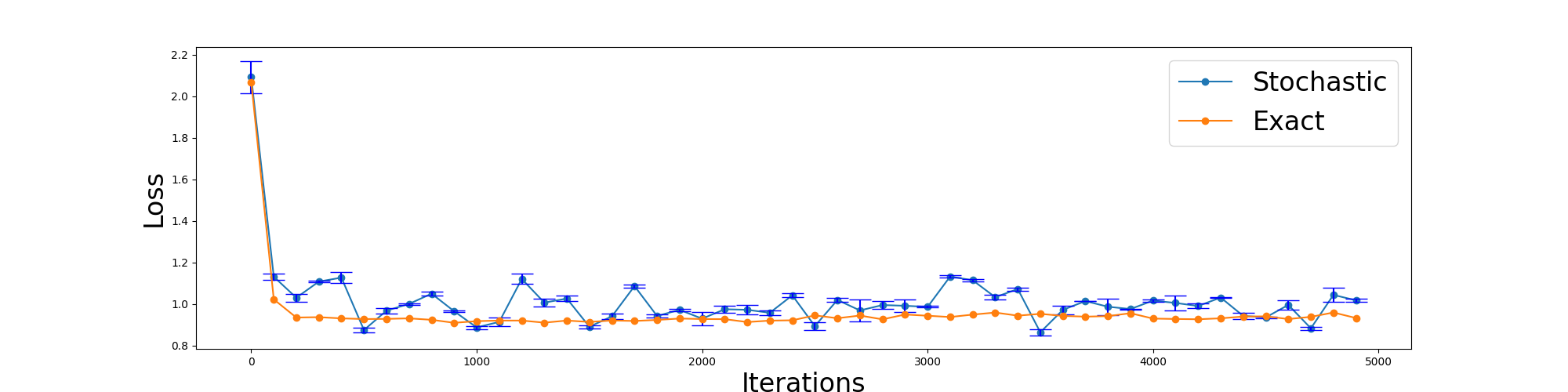}
    \caption{\textbf{Loss plot during training of the toy model.} Loss calculated with trace estimator successfully converges compared to that calculated with the exact trace value. Loss calculated with trace estimator was repeated 5 times.}
    \label{fig:toy_loss}
\end{figure*}

\subsection{Computational Comparison}
\label{appendix:ste:computation}
Given that $\mathcal{X} \subset \R^{256 \times 256 \times 3}$ and $\mathcal{H} \subset \R^{8 \times 8 \times 512}$, the encoder's Jacobian $\mJ$ contains 6,442,450,944 elements. With \texttt{float32} data type, the Jacobian matrix uses approximately 24 GB of memory.
The computation time for a single Jacobian takes 202.77 seconds under our environment using NVIDIA A100 40GB.

In contrast, the Jacobian Vector Product (JVP) does not explicitly calculate the entire Jacobian matrix, but it directly computes the product of the Jacobian matrix with a specific vector, requiring only ($256\times 256 \times 3 + 8 \times 8 \times 512) \times 4$ = 91,750 bytes, which is approximately 0.875MB of memory.
In our isometry loss, we utilize three times of JVPs for estimating the trace of a Jacobian.
The computation time for a single JVP takes 0.6 seconds under our environment.


\section{Illustration of the Isometric Loss}
\label{appendix:illustration}

\begin{figure}[ht]
  \captionsetup[subfigure]{labelformat=simple}
  \renewcommand\thesubfigure{}
  \centering
  \subfigure{\includegraphics[width=0.22\textwidth]{figures/s2_manifold.png}}
  \subfigure{\includegraphics[width=0.22\textwidth]{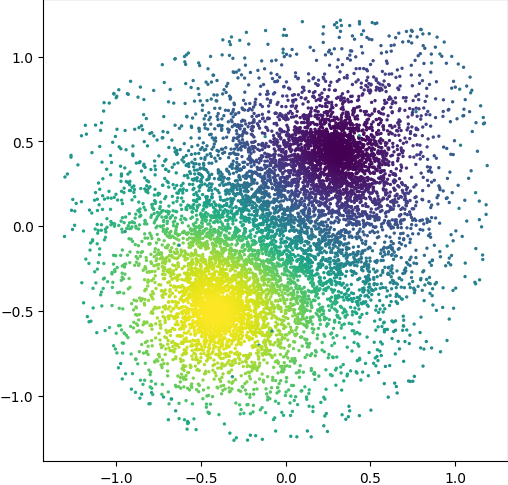}}
  \subfigure{\includegraphics[width=0.22\textwidth]{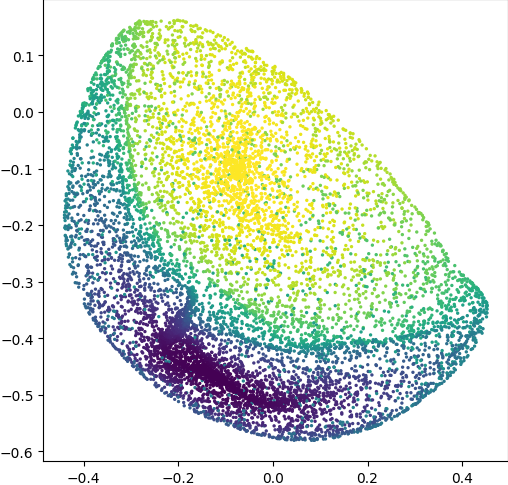}}
  \subfigure{\includegraphics[width=0.22\textwidth]{figures/isometric_G.png}} \\
  \subfigure[(a) $S^2$ manifold]
  {\includegraphics[width=0.22\textwidth]{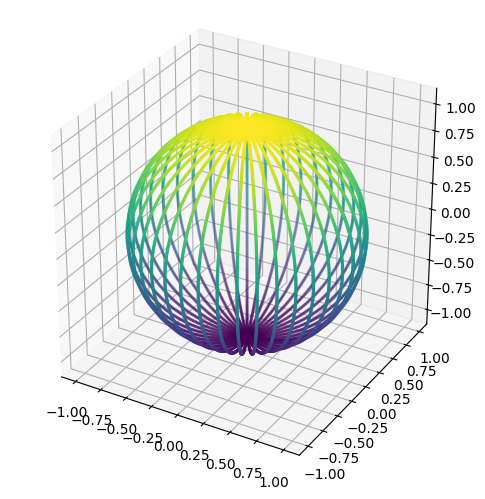}}
  \subfigure[(b) Reconstruction only]
  {\includegraphics[width=0.22\textwidth]{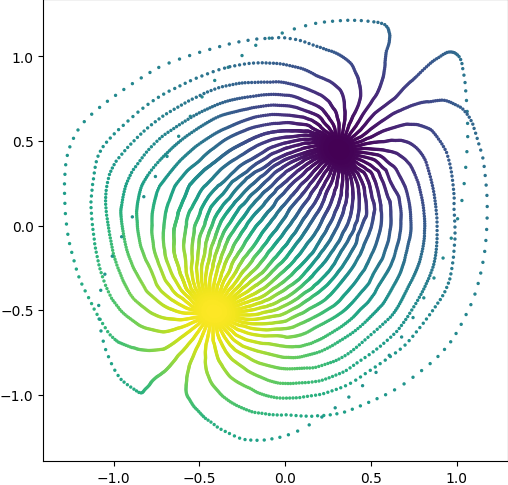}}
  \subfigure[(c) $\mathcal{L}_\text{iso}$ with $\mathbf{G} = \mI$ (Euclidean)]{\includegraphics[width=0.22\textwidth]{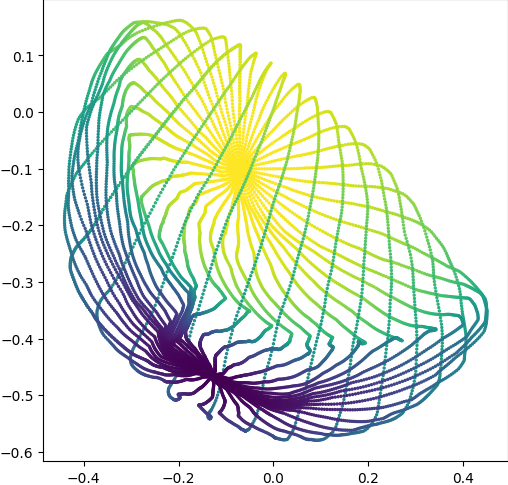}}
  \subfigure[(d) $\mathcal{L}_\text{iso}$ with full Eq. (\ref{eq:loss_iso})]
  {\includegraphics[width=0.22\textwidth]{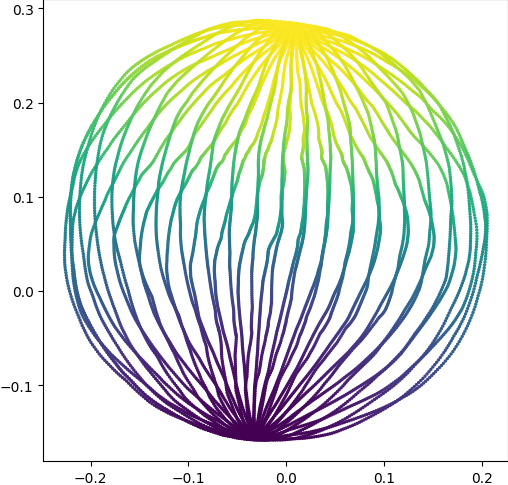}}
  \caption{(a) Illustration of the input $S^2$ manifold. (b--d) Mapped contours in latent coordinates learned by an autoencoder; (b) with reconstruction loss only, (c) with isometric loss assuming naive Euclidean geometry, and (d) with our isometric loss considering $S^2$ geometry.}
  \label{fig:more_toymodel}
\end{figure}

In Fig.~\ref{fig:more_toymodel}, we provide more illustrations of the latent space of an autoencoder, regularized with isometric loss.

\section{Preservation of $\mathcal{H}$ after Isometric Training}
\label{appendix:h_perservation}

{Trained with our isometric loss acting as a regularizer to the denoising score matching loss, it is not trivial if the model eventually learns the semantic space in $\mathcal{H}$.
However, \citet{kwon2023diffusion} argues that $\mathcal{H}$ exists in the bottleneck layer of the U-Net, for all pretrained diffusion models.
Hence, it is reasonable to deduce that $\mathcal{H}$ space exists given that the denoising score matching (DSM) loss has converged.
Therefore, it can be inferred that $\mathcal{H}$-space exists if the DSM loss converges to a similar point, even when the isometric loss is added.}

{We observe that the addition of the isometry loss does not significantly alter the convergence point of the diffusion loss and still shows comparable FID scores.
From this, we can naturally conclude that $\mathcal{H}$-space also still exists in our model.}

\begin{figure*}[ht]
    \centering    
    \includegraphics[width=\linewidth]{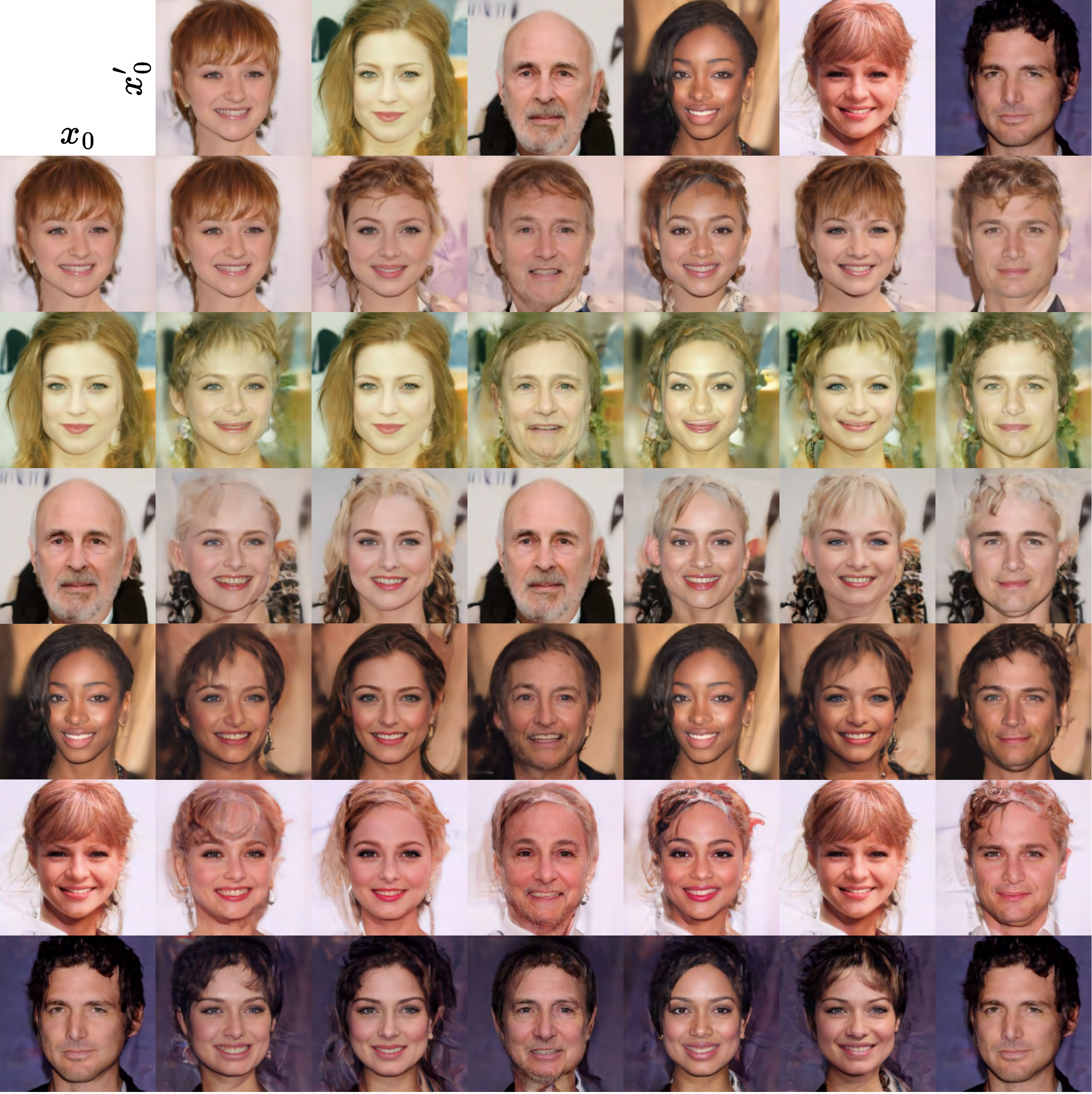}
    \caption{\textbf{Empirical observation regarding existence of $\mathcal{H}$ in our model.} Images in the same row share the original image $\vx_0$, images in the same column share the source image $\vx'_0$ for editing direction $\{h'_t\}$.}
    \label{fig:h_preservation}
\end{figure*}

{As empirical evidence, we provide some qualitative results of image editing with the  $\mathcal{H}$ in Fig.~\ref{fig:h_preservation}.
We aim to edit the image $\vx_0$ to the direction toward $\vx'_0$, manipulating only the content of the image while preserving the person's identity.
Specifically, we first calculate features $\{\vh_t\}$ and $\{\vh'_t\}$ corresponding to $\vx_t$ and $\vx'_t$, respectively, where $t$ is the DDIM time steps.
Then, we use $\{\vh_t + 1.5\vh'_t\}$ to inject contents during the reverse process starting from $\vx_T$, following \citet{jeong2023training}.
Note that the leftmost image for each row is $\vx_0$, and other images in the same row are the edited ones.}

\section{Latent Traversal Examples}
\label{appendix:interpolation}



\begin{figure*}[ht]
    \vspace{-0.3cm}
    \centering
    \includegraphics[width=\linewidth]{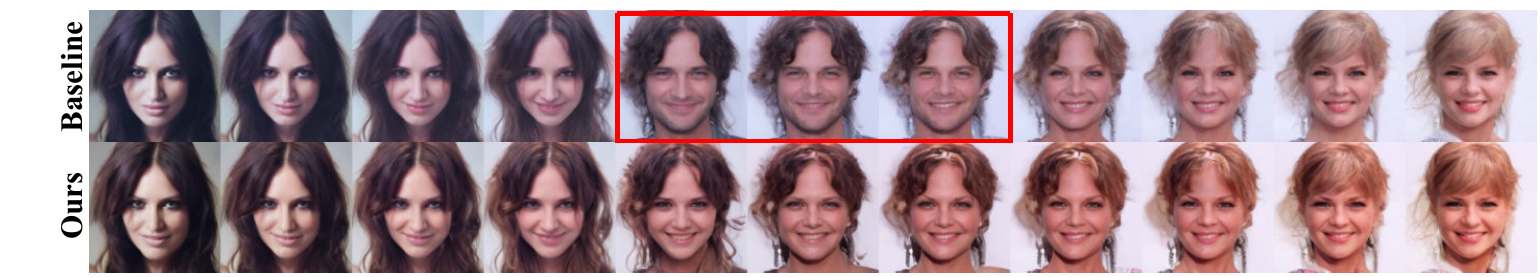}
    \includegraphics[width=\linewidth]{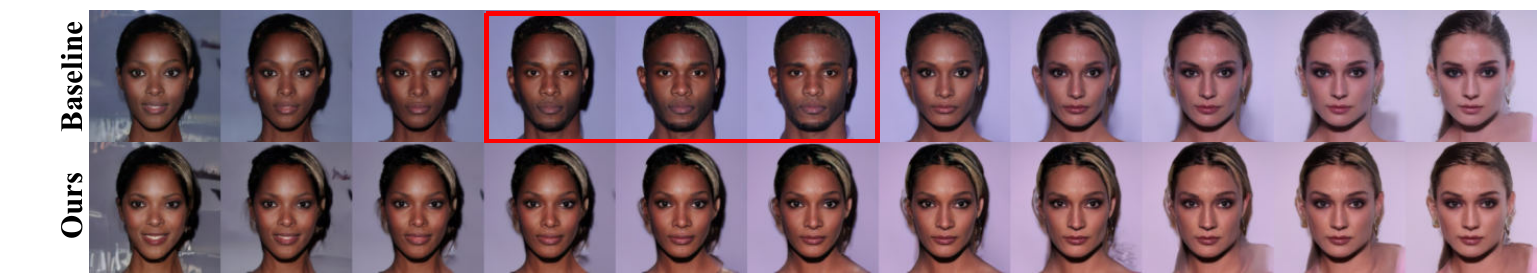}
    \includegraphics[width=\linewidth]{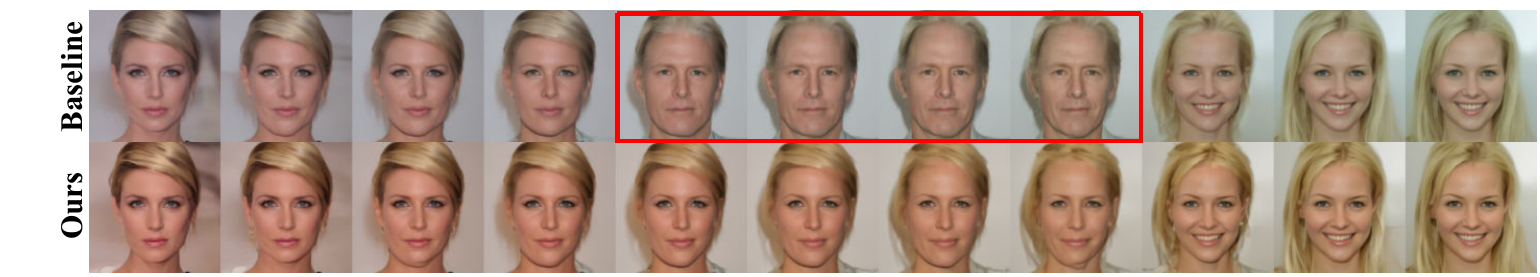}
    \includegraphics[width=\linewidth]{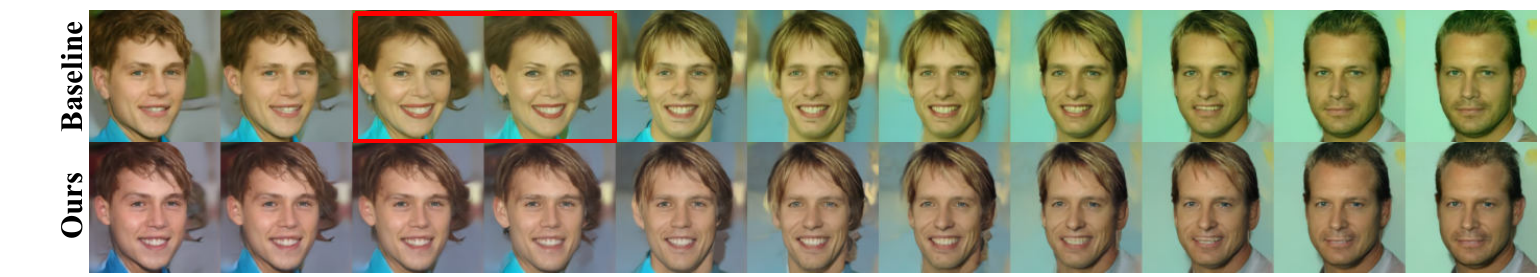}
    \includegraphics[width=\linewidth]{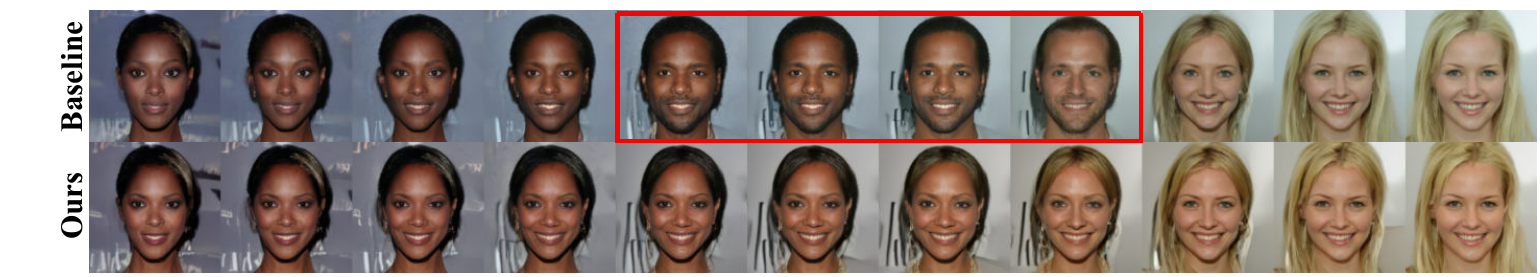}
    \includegraphics[width=\linewidth]{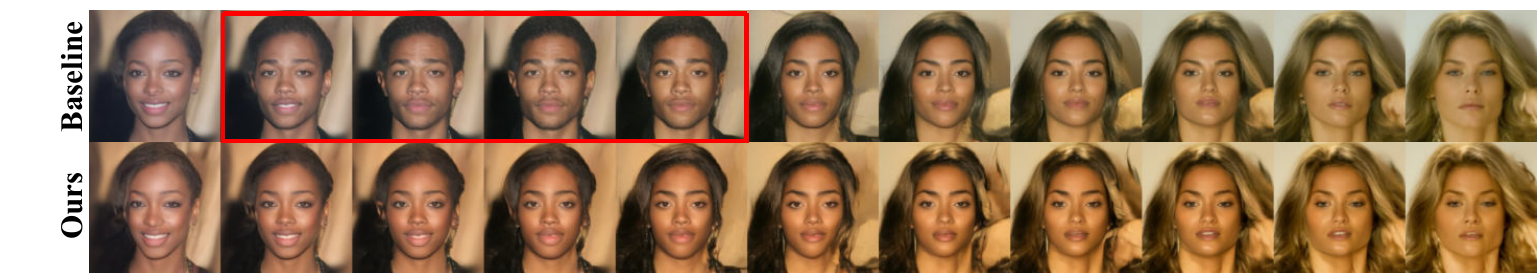}
    \caption{\textbf{Additional examples of latent traversal} between two images with DDPM and ours trained with isometric regularizer, trained on $256 \times 256$ CelebA-HQ.
    }
    \label{fig:interpolation2}
\end{figure*}

\begin{figure*}[ht]
    \vspace{-0.3cm}
    \centering
    \includegraphics[width=\linewidth]{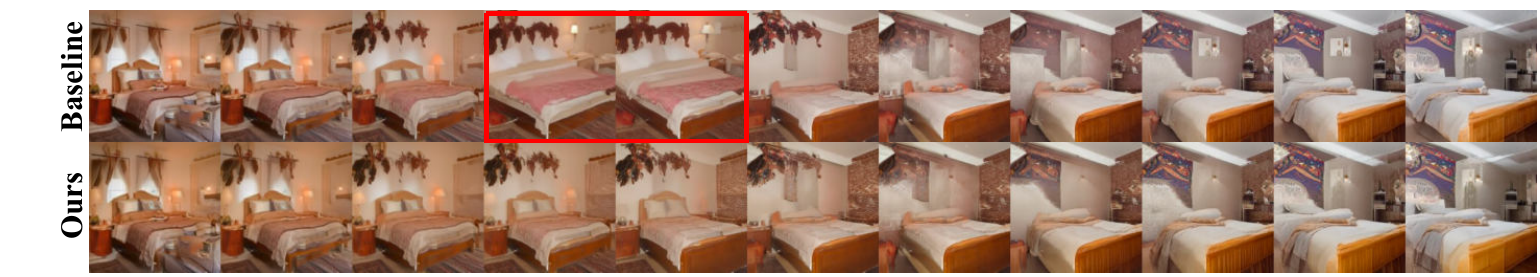}
    \includegraphics[width=\linewidth]{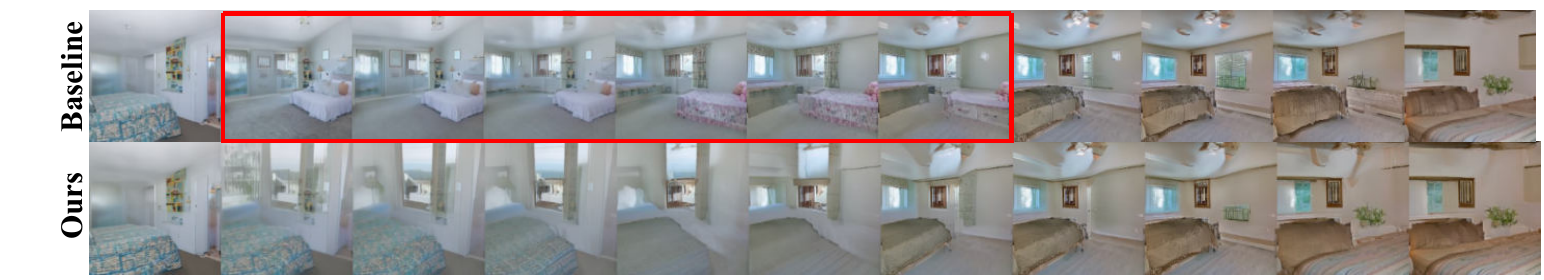}
    \includegraphics[width=\linewidth]{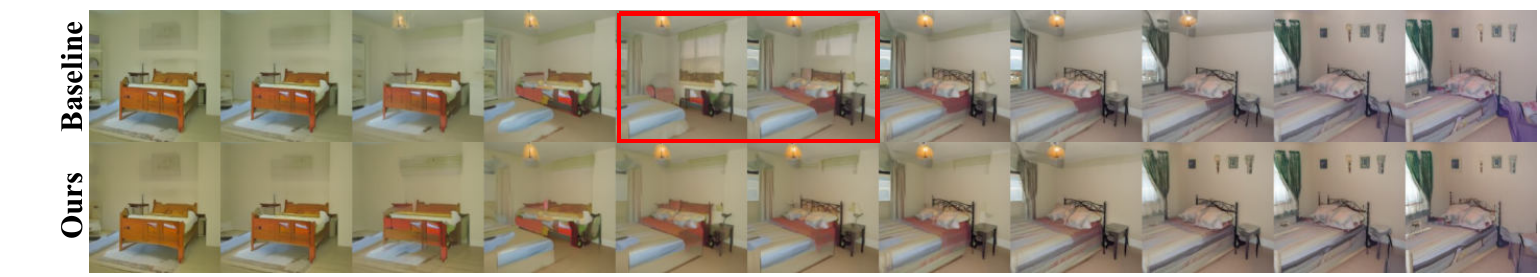}
    \includegraphics[width=\linewidth]{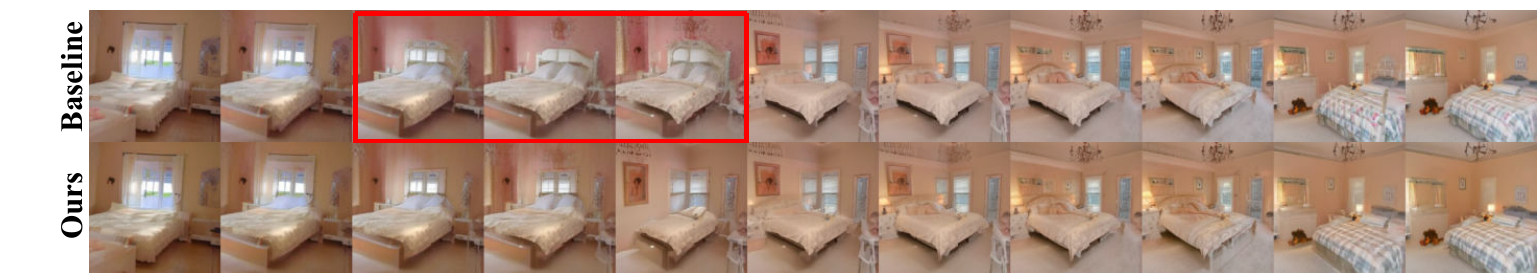}
    \includegraphics[width=\linewidth]{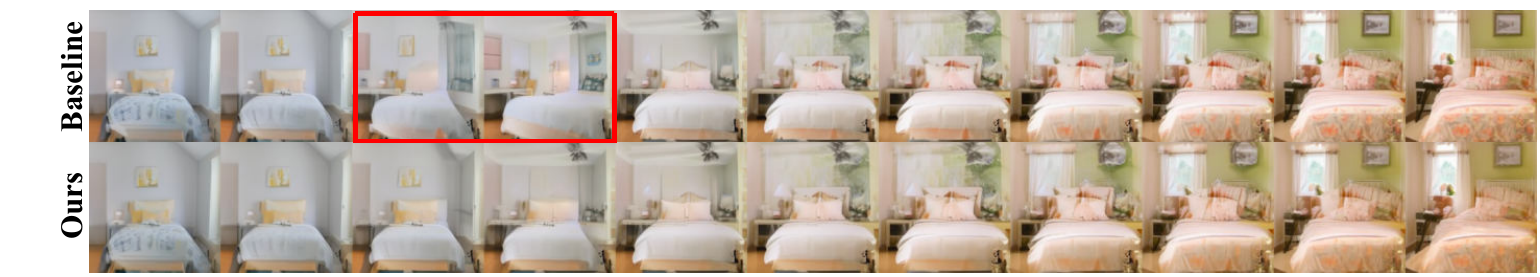} 
    \includegraphics[width=\linewidth]{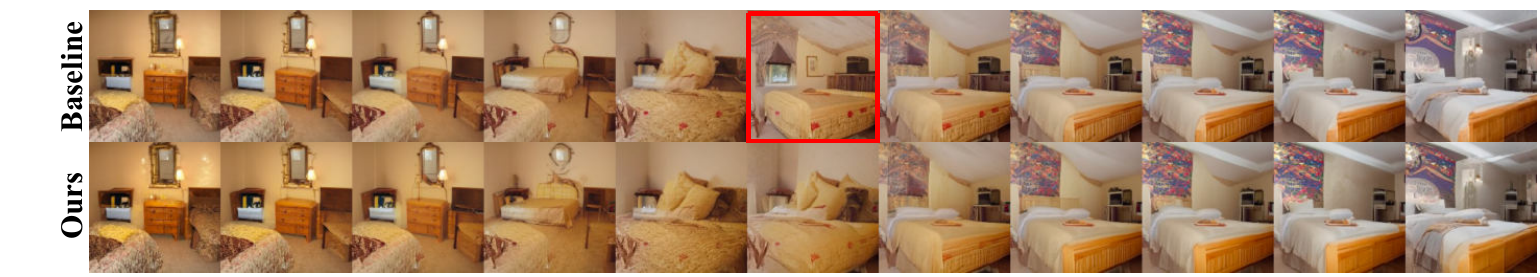}
    \caption{\textbf{Additional examples of latent traversal} between two images with DDPM and ours trained with isometric regularizer, trained on $256 \times 256$ LSUN-Bedroom.
    }
    \label{fig:interpolation3}
\end{figure*}

\begin{figure*}[ht]
    \vspace{-0.3cm}
    \centering
    \includegraphics[width=\linewidth]{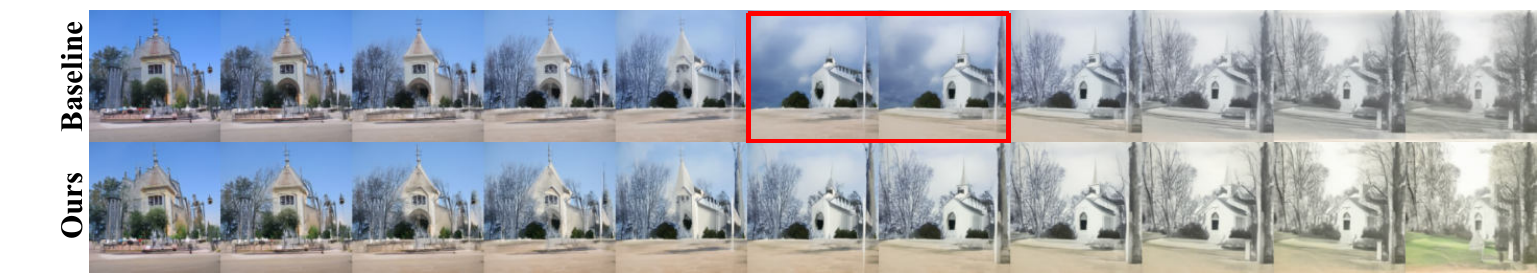}
    \includegraphics[width=\linewidth]{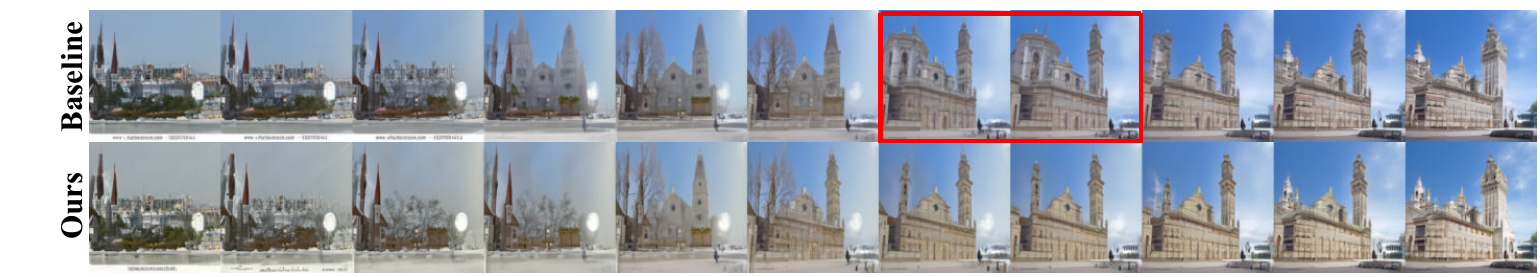}
    \includegraphics[width=\linewidth]{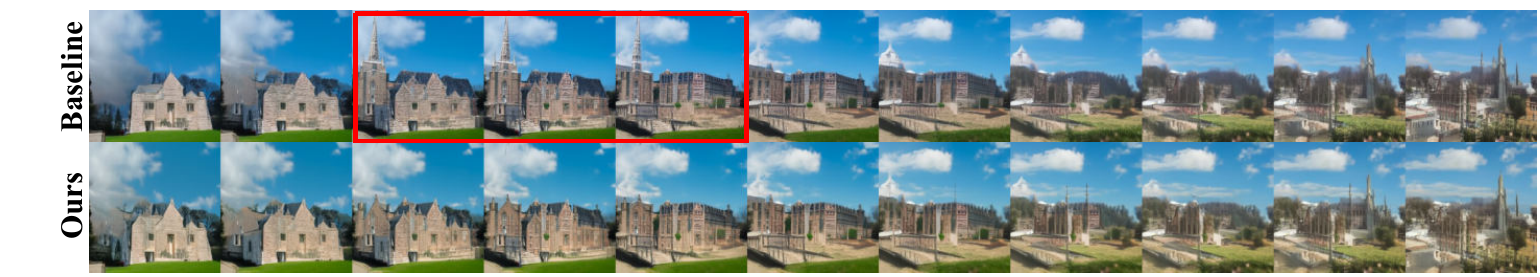}
    \includegraphics[width=\linewidth]{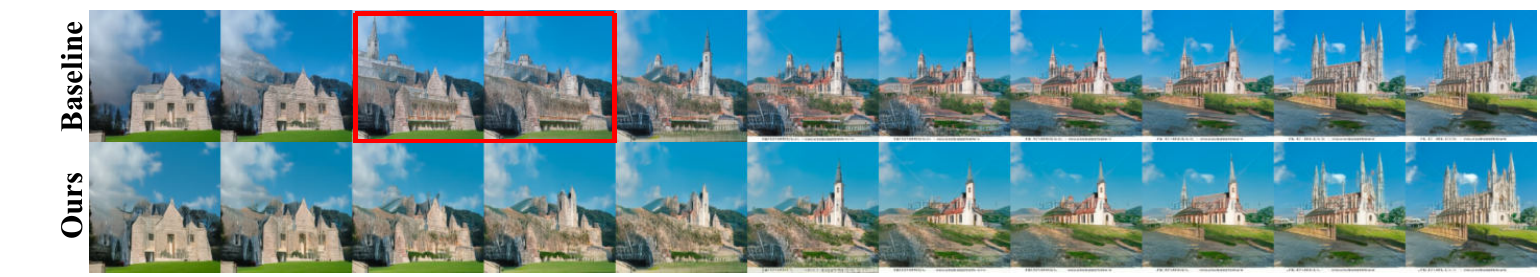}
    \includegraphics[width=\linewidth]{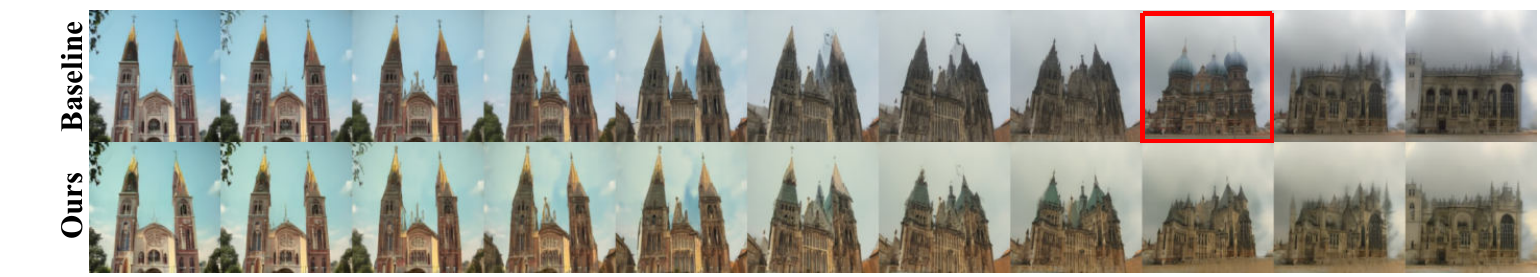} 
    \includegraphics[width=\linewidth]{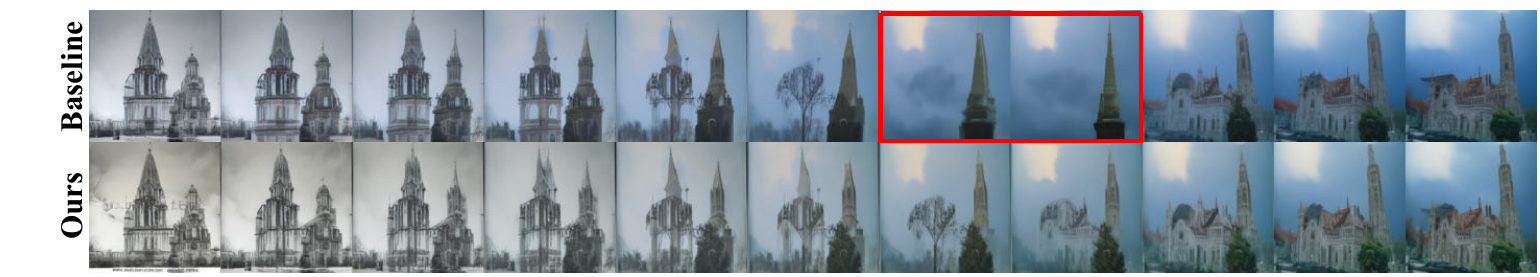}
    \caption{\textbf{Additional examples of latent traversal} between two images with DDPM and ours trained with isometric regularizer, trained on $256 \times 256$ LSUN-Church.
    }
    \label{fig:interpolation4}
\end{figure*}

We provide additional examples to compare the latent traversals with the baseline (DDPM) and with our model trained with isometric loss, trained on CelebA-HQ, LSUN-Bedroom, and LSUN-Church datasets.
The image resolution is $256 \times 256$ for all datasets.
Fig.~\ref{fig:interpolation2}--\ref{fig:interpolation4} extend Fig.~\ref{fig:interpolation} with more examples.

\newpage

\newpage

\newpage
